\documentclass[11pt]{article}

\usepackage[preprint]{acl}

\usepackage{times}
\usepackage{latexsym}
\usepackage{wrapfig}
\usepackage{array}             
\usepackage{tabularx}
\usepackage[most]{tcolorbox}   
\usepackage{xcolor}            
\usepackage{booktabs}          
\usepackage{array}
\usepackage{booktabs}
\usepackage{pgfplots}
\usepackage{amssymb}
\usepackage{wrapfig}
\usepackage{pgfplots}
\usepackage{booktabs}
\usepackage[table]{xcolor}
\usepackage{multirow}
\usepackage{graphicx}
\usepackage{booktabs}
\usepackage{multirow}
\usepackage[table]{xcolor}
\usepackage{graphicx}
\usepackage{booktabs}
\usepackage{multirow}
\usepackage[table]{xcolor}
\usepackage{graphicx}
\usepackage{array}

\newcommand{\modelrule}{%
  \specialrule{0.3pt}{0pt}{0pt}%
}
\newcommand{\groupline}[1]{%
  \noalign{\vskip-2.0pt}%
  \cmidrule(l{1pt}r{1pt}){#1}%
  \noalign{\vskip-2.0pt}%
}
\usepackage{forest}
\usepackage{wrapfig}
\usepackage[T1]{fontenc}

\usepackage[utf8]{inputenc}

\usepackage{microtype}

\usepackage{inconsolata}

\usepackage{graphicx}

\title{The Imperfective Paradox is Not Necessarily in Large Language Models:\\
A Benchmark Failure Before a Model Failure}

\author{
  Kaiqiao Han\textsuperscript{1} \quad
  Mark Steedman\textsuperscript{2} \quad
  Yizhou Sun\textsuperscript{1} \\
  \\
  \textsuperscript{1}University of California, Los Angeles \\
  \textsuperscript{2}University of Edinburgh \\
  \\
  \texttt{\{kqhan,yzsun\}@cs.ucla.edu} \\
  \texttt{\{m.steedman\}@ed.ac.uk}
}

\begin{document}
\maketitle
\begin{abstract}
Recent work uses an NLI benchmark of the imperfective paradox to argue that large language models exhibit a {Teleological Bias} and {Calibration Crisis}.
We identify three validity concerns in benchmark construction, including aspectual reduction, semantic, and relation bias.
Under a strict NLI standard, 76\% of the Group A instances do not explicitly rule out culmination; in native-speaker annotation, 38\% of Group A and 29\% of Group C examples permit an alternative interpretation.
To better isolate the semantic contrast, we construct {Lexically Matched Minimal Pairs}.
Moreover, by formulating event-semantic NLI as a {multi-step reasoning problem}, we evaluate intermediate semantic judgments and use oracle-guided settings. We find that incorrect NLI entailment does not fully support teleological bias.
Prompting often changes final NLI decisions without reliably improving the underlying semantic judgments, offering a possible explanation for the {Calibration Crisis}.

\end{abstract}
\section{Introduction}

Substantial evidence suggests that smaller language models often rely on surface-level statistical cues rather than systematic compositional analysis in tasks requiring semantic composition, especially under unguided, zero-shot, direct-answer settings without CoT~\citep{gururangan-etal-2018-annotation,mccoy-etal-2019-right,kim-linzen-2020-cogs, qiu-etal-2022-evaluating,wei2023chainofthoughtpromptingelicitsreasoning}. A recent testbed for this limitation is the Imperfective Paradox.
It captures a contrast between telic predicates, for which an ongoing event does not entail culmination, and atelic predicates, for which the corresponding progressive can support a simple-past inference. Box~\ref{box:imperfective_paradox} illustrates this contrast~\citep{vallurupalli,prus-etal-2025-world,imperfective}. 
Recent work argues that LLMs exhibit a \textbf{Teleological Bias}, tending to infer that goal-directed events reach their intended endpoints. Using an NLI benchmark contrasting ``telic and atelic verbs,'' it reports that even state-of-the-art open-source models\footnote{Mainly 7--9B models, such as Qwen2.5-7B-Instruct.} frequently infer completion for interrupted telic events. Prompting interventions further trigger a \textbf{Calibration Crisis}: strategies such as Chain-of-Thought or forced counterfactual non-endpoints can reduce hallucinated completions but often over-correct by rejecting valid atelic entailments, supporting the claim that models struggle to adapt their reasoning to event type~\citep{imperfective}.

\begin{tcolorbox}[
title=\textbf{The Imperfective Paradox},
colback=gray!5,
colframe=gray!70,
boxrule=0.5pt,
arc=2mm,
left=2mm,
right=2mm,
top=1mm,
bottom=1mm]
\label{box:imperfective_paradox}
\small

\textbf{Telic Predicate}
\hfill
\makebox[3.2cm][l]{\textbf{Entailment:} \textcolor{red}{Unknown}}

\textit{Premise}: The carpenter was building a gazebo.\\
\textit{Hypothesis}: The carpenter built a gazebo.

\vspace{0.8ex}
\hrule
\vspace{0.8ex}

\textbf{Atelic Predicate}
\hfill
\makebox[3.2cm][l]{\textbf{Entailment:} \textcolor{green!60!black}{True}}

\textit{Premise}: The boy was running in the park.\\
\textit{Hypothesis}: The boy ran in the park.

\end{tcolorbox}

In this paper, we argue that both the benchmark design and the conclusions drawn from it require reexamination. We organize our analysis around three questions:
\textbf{(1) Does the benchmark isolate the intended imperfective-paradox contrast?}
\textbf{(2) Do final NLI errors and prompting-induced shifts provide valid evidence for the claimed Teleological Bias and Calibration Crisis?}
and
\textbf{(3) Once the task is controlled, what limits models' event-semantic reasoning?}
Together, these questions separate three increasingly stronger claims: whether the benchmark measures the intended semantic construct, whether observed prediction errors reliably diagnose a particular semantic failure, and which components of event-semantic reasoning remain challenging when these factors are examined separately.

A central issue in benchmark validity is \textbf{Aspectual Reduction}: the benchmark assigns aspectual classes at the verb level, although aspect is a compositional property of the complete predicate. This issue is particularly consequential because these verb-level classifications determine the benchmark grouping and serve as the basis for the subsequent experimental design and analysis.
We additionally identify two concerns. First, the benchmark is affected by \textbf{{Semantic Bias}}, treating culmination as a binary and context-independent condition. Some predicates also permit both telic and atelic interpretations. Second, the benchmark is affected by \textbf{Relation Bias} that conflate interruption with non-completion.
Under a strict NLI standard, 76\% of Group A instances do not explicitly establish non-completion, and in our native-speaker annotation, {38\%} of Group A examples and {29\%} of the Group C examples were judged to permit an alternative interpretation. Furthermore, the original benchmark confounds compositional aspect with {uncontrolled lexical variation}. 
We therefore construct
\textbf{Lexically Matched Minimal Pairs} that provide a controlled
evaluation. 

Beyond benchmark validity, a mismatch between model predictions and benchmark labels does not by itself identify the semantic judgment responsible for it. Event-semantic NLI requires several distinct decisions, yet existing evaluations collapse them into a single classification step. We therefore formulate the task as a \textbf{Multi-Step Reasoning Framework} and evaluate both intermediate semantic judgments and final predictions.
The characteristic NLI error previously attributed to Teleological Bias frequently occurs {without} a corresponding culmination error. Models often judge culmination as unsupported or unresolved while still accepting the corresponding telic simple-past hypothesis. Our analysis further shows that models may treat evidence of event occurrence as sufficient for accepting a telic predicate without enforcing its stronger culmination requirement, a behavioral inconsistency we term \textbf{Sufficiency Error}. Thus, the same final NLI error can arise from different failures and does not uniquely diagnose Teleological Bias.
We further find that prompting can substantially shift final NLI predictions without corresponding improvements in the intermediate semantic judgments required by the task, a phenomenon we term \textbf{Decision Shift}. The reported Calibration Crisis may therefore partly reflect changes in downstream decision behavior rather than changes in the underlying event-semantic analysis.

Intermediate-output and
oracle-guided evaluations reveal further failures beyond the
culmination decision itself. In particular, models frequently misclassify the aspectual type of the {complete predicate}. Errors also arise when mapping the resulting semantic relation to the final NLI label.
Results from Qwen2.5-72B and GPT-5.4 provide evidence that aspectual classification can be context-sensitive even in stronger models, and models can correctly identify non-culmination yet still accept the simple-past form of a telic predicate. 


Our contributions are summarized as follows:
(1)
We identify three recurring {Benchmark Validity Concerns }in the benchmark. We construct {Lexically Matched Minimal Pairs} for more controlled evaluation.
(2) 
We show that characteristic NLI errors often occur without corresponding culmination errors, undermining a reliable diagnostic of Teleological Bias. Some prompts can likewise shift final decisions without improving the underlying semantic reasoning.
(3)
We formulate event-semantic NLI as a{ multi-step reasoning framework} and use intermediate-output and oracle-guided evaluations to analyze errors in compositional aspect classification and semantic-to-label mapping.

\section{Related Work}

\subsection{Aspectual Classes and Interpretive Variation}

Vendler distinguishes four aspectual classes based on the temporal properties of
predicates. Crucially, aspectual class is compositionally determined by the
complete predicate rather than the isolated verb: contrasts such as
\textit{run} versus \textit{run a mile} arise from the interaction of the verb
with its arguments and other sentential material
~\citep{1f85e7b1-f91a-3736-a7b0-868da43ec6eb,filip1999aspect}.
Human judgments also show substantial variation, particularly for telic
predicates, even under explicit interruption contexts. Among 30 theoretically
telic predicates, only 4 were strongly judged as showing the expected
progressive-to-completed non-entailment; participants instead predominantly
accepted the completed-event inference for 2 of 10 telic-punctual predicates
and 12 of 20 telic-durative predicates
~\citep{prus-etal-2025-world,prus-etal-2024-human}. 

\subsection{Imperfective Paradox Benchmarks}

CoRE evaluates the imperfective paradox by comparing model judgments of
ongoing and completed event descriptions with and without narrative context.
Its results show that additional context does not consistently improve model
performance, suggesting limited sensitivity to aspectual distinctions
~\citep{vallurupalli}.
The benchmark studied in this work adopts a $2\times2$ design crossing
predicate telicity with contextual outcome information
~\citep{imperfective}. Telic accomplishments and atelic activities are paired
with either an interruption context or an outcome-underspecified progressive
context, yielding four groups: interrupted accomplishments (Group A),
interrupted activities (Group B), ambiguous accomplishments (Group C), and
ambiguous activities (Group D), as illustrated in
Table~\ref{tab:imperfective_benchmark}.






















\begin{table}[t]
\centering

\scriptsize
\setlength{\tabcolsep}{4pt}
\renewcommand{\arraystretch}{1.12}

\begin{tabularx}{\columnwidth}{
@{}>{\raggedright\arraybackslash}p{0.19\columnwidth}
>{\raggedright\arraybackslash}X
>{\raggedright\arraybackslash}X@{}
}
\toprule
&
\textbf{Telic}
&
\textbf{Atelic}
\\
\midrule

\textbf{Interruption}
&
\textbf{Group A (\textsc{False})}

\textit{The carpenter was building a gazebo, but a storm destroyed the frame.}

\textbf{H:} The carpenter built a gazebo.
&
\textbf{Group B (\textsc{True})}

\textit{The athletes were running on the track, but it started to hail.}

\textbf{H:} The athletes ran on the track.
\\
\midrule

\textbf{Outcome unspecified}
&
\textbf{Group C (\textsc{Unknown})}

\textit{The carpenter was building a gazebo.}

\textbf{H:} The carpenter built a gazebo.
&
\textbf{Group D (\textsc{True})}

\textit{The athletes were running on the track.}

\textbf{H:} The athletes ran on the track.
\\

\bottomrule
\end{tabularx}
\caption{
The Imperfective Paradox benchmark.
}
\label{tab:imperfective_benchmark}

\end{table}
\section{Does the Benchmark Isolate the Intended Contrast?}
\label{sec:benchmark_issues}

\begin{tcolorbox}[
title=\textbf{Benchmark Bias},
colback=gray!4,
colframe=gray!65,
boxrule=0.5pt,
arc=2mm,
left=1.5mm,
right=1.5mm,
top=1mm,
bottom=1mm
]

\small

\textbf{(1) Aspectual Reduction: Verb $\neq$ Predicate}

\textit{run in the park} is atelic, whereas
\textit{run home} can be telic.
Aspect therefore cannot be inherited from the lexical verb alone.

\vspace{0.7em}

\textbf{(2a) Completion need not require a maximal endpoint}

\textit{John ate an apple, leaving only a tiny piece.}
A contextually negligible remainder need not invalidate an ordinary
completion judgment.

\vspace{0.7em}

\textbf{(2b) Predicate construal may vary}

\textit{She was climbing the mountain.}
The predicate may receive a bounded endpoint-oriented or a less strongly
resultative interpretation.

\vspace{0.7em}

\textbf{(3) Interruption $\neq$ Non-completion}

\textit{Construction stopped for a day.}
Interruption alone establishes neither eventual completion nor
non-completion; the corresponding NLI relation may therefore be
\textsc{Neutral} rather than \textsc{Contradiction}.

\end{tcolorbox}

 We identify three sources of construction bias:
(1)\textbf{ Aspectual Reduction}: aspectual classes are assigned using verb-level heuristics rather than predicate-level composition;
(2) \textbf{Semantic Bias}: event completion is defined using an overly rigid canonical-endpoint criterion or some predicates permit both atelic and telic
uses;
(3) \textbf{Relation Bias}: interruption is mapped to contradiction in the NLI annotations.
These concerns systematically affect how model behavior is evaluated.

\subsection{Aspectual Reduction}
\label{sec:predicate_aspect}

The first concern is \textbf{Aspectual Reduction}: the aspectual class is treated
as a fixed property of an isolated verb, although it is compositionally
determined by the complete predicate, including its arguments, path
expressions, and quantization properties. For example, \textit{run in the
park} is typically atelic, whereas \textit{run home} or \textit{run a mile}
can be telic because they introduce a bounded path or extent
~\citep{filip1999aspect}. Benchmark instance \texttt{A\_098} illustrates this issue:

\begin{quote}
\small
\textbf{Premise:} We were picking strawberries, but a bear chased us off.\\
\textbf{Hypothesis:} We picked strawberries.\\
\textbf{Original label:} \textsc{False}
\end{quote}


\subsection{Semantic Bias}
\label{sec:completion_semantics}

The second concern is the benchmark's treatment of completion and predicate
interpretation. It often assumes that a telic event counts as completed only
when a canonical endpoint is fully realized, although completion judgments
may depend on lexical semantics and contextual standards
~\citep{filip1999aspect,filip2008}. Moreover, some predicates admit both
more strongly bounded and more activity-like construals, while the benchmark
assigns a single categorical interpretation
~\citep{filip1999aspect,filip2008,article}. For example, benchmark instance \texttt{C\_018} illustrates this issue:

\begin{quote}
\small
\textbf{Premise:} He was cleaning the garage.\\
\textbf{Hypothesis:} He cleaned the garage.\\
\textbf{Original label:} \textsc{Unknown}
\end{quote}


\subsection{Relation Bias}
\label{sec:interruption_nli}

The third concern arises when event information is mapped to NLI labels. The
benchmark frequently treats interruption as evidence that the corresponding
completed-event statement is false.
However, interruption alone does not establish non-completion and may therefore license \textsc{Neutral}, rather than \textsc{Contradiction}. An interrupted event may already count as completed, may resume and later culminate, or may simply have an outcome that remains unspecified. For
example, benchmark instance \texttt{A\_081} makes this distinction 
visible~\citep{bowman-etal-2015-large,3ce5ded6b8774adfb7c7eb541ebc6e98}:

\begin{quote}
\small
\textbf{Premise:} She was cutting the fabric for the dress, but realized she
cut it wrong.\\
\textbf{Hypothesis:} She cut the fabric for the dress.\\
\textbf{Original label:} \textsc{False}
\end{quote}



\begin{figure}[t]
\centering

\begin{tikzpicture}
\begin{axis}[
    width=\columnwidth,
    height=4.3cm,
    xlabel={Model Size},
    ylabel={Accuracy},
    symbolic x coords={1.5B,7B,14B,32B,72B},
    xtick=data,
    ymin=0,
    ymax=1.05,
    ytick={0,0.2,0.4,0.6,0.8,1.0},
    grid=major,
    major grid style={dashed},
    tick label style={font=\footnotesize},
    label style={font=\small},
    legend style={
        font=\footnotesize,
        at={(0.02,0.98)},
        anchor=north west,
        draw=none,
        fill=none
    },
    mark size=2pt,
    line width=0.9pt
]

\addplot[
    mark=*,
] coordinates {
    (1.5B,0.21)
    (7B,0.20)
    (14B,0.24)
    (32B,0.53)
    (72B,0.43)
};
\addlegendentry{Group A}

\addplot[
    mark=square*,
] coordinates {
    (1.5B,0.00)
    (7B,0.47)
    (14B,0.39)
    (32B,0.91)
    (72B,0.84)
};
\addlegendentry{Group C}

\end{axis}
\end{tikzpicture}

\caption{Accuracy on Groups A and C across model sizes
~\citep{imperfective}.}
\label{fig:accuracy-ac-model-size}
\end{figure}
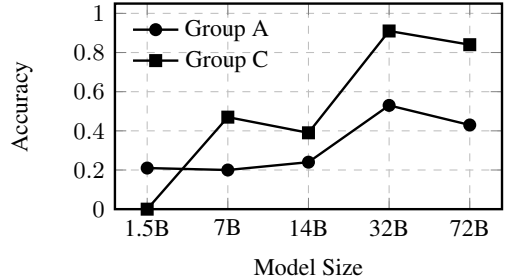

Figure~\ref{fig:accuracy-ac-model-size} provides further evidence for this
distinction. Accuracy on Group C improves sharply with model size, whereas
Group A remains substantially lower and less stable. This asymmetry is
consistent with a label-level conflict in Group A: models may recognize that
interruption leaves the eventual outcome unresolved and predict
\textsc{Neutral}, while the benchmark assigns \textsc{Contradiction}. Thus,
some apparent completion errors may instead arise from relation bias.

Taken together, these cases suggest that the intended semantic distinctions are not always cleanly reflected in the benchmark labels. 




\subsection{Controlling Semantic Confounds with Lexically Matched Minimal Pairs}
\label{sec:minimal_pairs}

A further limitation of the original benchmark is insufficient control over lexical and contextual variation: activity and accomplishment examples often differ in verbs, participants, objects, and scenarios. As a result, prediction differences may reflect lexical associations, contextual plausibility, or alternative event interpretations rather than aspectual reasoning alone.
To better isolate the intended semantic contrasts, we construct \textbf{lexically matched minimal pairs} in a controlled $2\times2$ design as shown in Table~\ref{tab:minimal_pair_construction}. Each four-example set shares the same subject, lexical verb, tense alternation, and general scenario, while varying only (1) {predicate telicity}, whether the complete predicate is telic or atelic, and (2) {outcome evidence}, whether non-completion is explicitly established or left unspecified. Crucially, telicity is manipulated compositionally at the predicate level rather than through the lexical verb.

\paragraph{Implementation Details} In the following sections, Original (Orig.) denotes the original dataset, Validity-Screened Subset (VSS) denotes the dataset obtained after filtering the partial instances from original datasets, and Minimal Pairs (Min.) denotes the lexically matched dataset. We report results on the original benchmark and the independently constructed minimal pairs, with the VSS included as an auxiliary analysis. For Group A, we report the Unknown and False rates both separately and jointly to provide a more comprehensive analysis. In the main experimental tables, we report the joint rate for the original dataset and the False rate for the minimal-pair dataset, while the Appendix provides all detailed results, including both the separate and joint rates.
For implementation details, see Appendix~\ref{de}. For models, we use four open-source language models of comparable scale: Llama-3.1-8B (denoted as {Llama}), Qwen2.5-7B ({Qwen}), GLM-4-9B ({GLM}), and DeepSeek-R1-7B ({DeepSeek}).

\begin{table*}[t]
\centering

\small
\setlength{\tabcolsep}{8pt}
\renewcommand{\arraystretch}{1.18}

\begin{tabularx}{\textwidth}{
@{}>{\raggedright\arraybackslash}p{0.15\textwidth}
>{\raggedright\arraybackslash}X
>{\raggedright\arraybackslash}X@{}
}
\toprule
&
\textbf{Telic: walk \textbf{to} the station}
&
\textbf{Atelic: walk \textbf{in} the station}
\\
\midrule

\textbf{Explicit outcome}
&
\textbf{Group A (\textsc{Contradiction})}

\textbf{P:} Maya was walking \textbf{to} the station,
but stopped halfway and \textbf{never resumed}.

\textbf{H:} Maya walked \textbf{to} the station.
&
\textbf{Group B (\textsc{Entailment})}

\textbf{P:} Maya was walking \textbf{in} the station,
but stopped and never resumed.

\textbf{H:} Maya walked \textbf{in} the station.
\\
\midrule

\textbf{Outcome unspecified}
&
\textbf{Group C (\textsc{Neutral})}

\textbf{P:} Maya was walking \textbf{to} the station.

\textbf{H:} Maya walked \textbf{to} the station.
&
\textbf{Group D (\textsc{Entailment})}

\textbf{P:} Maya was walking \textbf{in} the station.

\textbf{H:} Maya walked \textbf{in} the station.
\\
\bottomrule
\end{tabularx}

\vspace{1mm}
\footnotesize
\textit{Shared lexical frame:}
Subject = \textit{Maya}; verb = \textit{walk};
progressive/simple-past alternation is fixed.
\caption{
Lexically matched minimal-pair construction.
All examples share the same subject and verb, while the telic/atelic contrast
is induced by the prepositional phrase, and the progressive/simple-past
alternation is held fixed.
}
\label{tab:minimal_pair_construction}
\end{table*}

\section{Do Final NLI Errors Support the Claimed Bias?}
In this section, we first clarify the evaluation bias and then present a deeper diagnostic analysis in the next section.

\subsection{Event-Semantic NLI as a Multi-Step Reasoning Process}
\label{sec:reasoning_decomposition}

We formulate event-semantic NLI as a multi-step reasoning problem. Correctly solving an example involves the eight reasoning steps shown in Table~\ref{tab:reasoning-steps}. We operationalize these components under different reasoning-prompt settings, which specify or elicit different subsets of the reasoning process.
The complete reasoning chain can be summarized as in Table~\ref{tab:reasoning-steps}.
These steps correspond to distinct linguistic and logical decisions. A final classification error therefore does not necessarily imply a failure to represent event completion.
\begin{table*}[t]
\centering

\small
\setlength{\tabcolsep}{6pt}
\renewcommand{\arraystretch}{1.12}

\begin{tabularx}{\textwidth}{
@{}c
>{\raggedright\arraybackslash}p{0.20\textwidth}
>{\raggedright\arraybackslash}X
>{\raggedright\arraybackslash}p{0.23\textwidth}@{}
}
\toprule
\textbf{Step}
& \textbf{Component}
& \textbf{Diagnostic Question}
& \textbf{Example Output} \\
\midrule

1 & Predicate Identification
& What is the complete event predicate?
& \textit{build a house} \\

2 & Aspectual Classification
& Is the complete predicate telic or atelic?
& Telic \\

3 & Process Verification
& Does the premise establish that the event process occurred?
& Yes \\

4 & Endpoint Identification
& What endpoint, if any, is encoded by the predicate?
& The house is completed \\

5 & Culmination Verification
& Does the premise entail that the endpoint was reached?
& Not entailed \\

6 & Non-Culmination Verification
& Does the premise entail that the endpoint was not reached?
& Yes / No \\

7 & Semantic Rule Application
& What relation follows from the semantic judgments above?
& Contradiction \\

8 & NLI Label Alignment
& How is that relation mapped to the benchmark label?
& \textsc{False} \\

\bottomrule
\end{tabularx}
\caption{
Multi-step diagnostic procedure for event-semantic NLI.
}
\label{tab:reasoning-steps}
\end{table*}

\subsection{Do Final Errors Necessarily Reflect Teleological Bias?}

Prior work attributes errors on telic predicates to \textbf{Teleological Bias},
arguing that models systematically infer culmination even when it is not
established by the premise~\citep{imperfective}. However, the same final NLI
error can arise even when the model does not represent the event as completed.
We first directly probe whether models judge the endpoint of the relevant
predicate to be reached. As shown in Table~\ref{tab:completion-group-results},
models frequently reject or remain uncertain about culmination while still
accepting the corresponding telic simple-past hypothesis. This suggests that
the failure need not lie in the culmination judgment itself. Instead, it may
arise when an unresolved semantic state is converted into an NLI decision:
evidence that an event occurred may be mapped to \textsc{Entailment}, even
though the absence of guaranteed culmination should leave the hypothesis
\textsc{Neutral}.
To further analyze this behavior, we construct an occurrence-to-completion
experiment with progressively stronger evidence: intention, occurrence,
interruption, explicit non-completion, and completion. The results in
Table~\ref{tab:occurrence-experiment} show that the largest change in model
decisions often occurs once \emph{event occurrence} is established, rather
than when culmination is established. In other words, models may correctly
recognize that an event took place while failing to enforce the stronger
culmination requirement associated with a telic simple-past predicate.
The direct state comparisons in Table~\ref{tab:completion-group-results}
provide converging evidence. Models often characterize the premise as
describing an unresolved or interrupted event rather than a completed one,
yet may still classify the corresponding telic hypothesis as entailed. Thus,
the characteristic final error does not uniquely diagnose an erroneous
culmination inference.
Taken together, these results reveal a behavioral inconsistency, which we
call \textbf{Sufficiency  Error}: models treat evidence of event occurrence as
sufficient for accepting a telic simple-past predicate, even when culmination
remains unsupported or unresolved. 
\begin{table}[t]
\centering

\scriptsize
\setlength{\tabcolsep}{2.0pt}
\renewcommand{\arraystretch}{1.06}

\resizebox{\columnwidth}{!}{%
\begin{tabular}{
@{}ll
c>{\columncolor{gray!12}}c
@{\hspace{5pt}}
c>{\columncolor{gray!12}}c
@{\hspace{4pt}}
c>{\columncolor{gray!12}}c
@{\hspace{4pt}}
c>{\columncolor{gray!12}}c
@{\hspace{4pt}}
c>{\columncolor{gray!12}}c
@{}
}
\toprule

\rowcolor{white}
\multirow{3}{*}{\textbf{Model}}
& \multirow{3}{*}{\textbf{Prompt}}
& \multicolumn{2}{c}{\multirow{2}{*}{\textbf{Completion}}}
& \multicolumn{8}{c}{\textbf{Group-wise Results}} \\

\cmidrule(lr){5-12}

\rowcolor{white}
&
&
&
& \multicolumn{2}{c}{\textbf{A}}
& \multicolumn{2}{c}{\textbf{B}}
& \multicolumn{2}{c}{\textbf{C}}
& \multicolumn{2}{c}{\textbf{D}} \\

\cmidrule(lr){3-4}
\cmidrule(lr){5-6}
\cmidrule(lr){7-8}
\cmidrule(lr){9-10}
\cmidrule(lr){11-12}

\rowcolor{white}
&
& \textbf{Acc.} & \textbf{NT}
& \textbf{Acc.} & \textbf{NC}
& \textbf{Acc.} & \textbf{NC}
& \textbf{Acc.} & \textbf{NC}
& \textbf{Acc.} & \textbf{NC} \\

\midrule

\multirow{2}{*}{Llama}
& NR
& 0   & 100
& 0   & 100
& 0   & 100
& 90  & 90
& 100 & 100 \\

& R
& 85 & 95
& 75 & 100
& 60 & 95
& 85 & 85
& 85 & 85 \\

\modelrule

\multirow{2}{*}{Qwen}
& NR
& 45 & 100
& 70 & 85
& 0  & 40
& 65 & 65
& 10 & 10 \\

& R
& 100 & 100
& 85  & 100
& 45  & 100
& 100 & 100
& 100 & 100 \\

\modelrule

\multirow{2}{*}{GLM}
& NR
& 0   & 100
& 0   & 70
& 0   & 100
& 100 & 100
& 100 & 100 \\

& R
& 80 & 100
& 55   & 100
& 15   & 100
& 100  & 100
& 100  & 100 \\

\modelrule

\multirow{2}{*}{DeepSeek}
& NR
& 55  & 100
& 65  & 100
& 30  & 90
& 100 & 100
& 100 & 100 \\

& R
& 90  & 100
& 80  & 100
& 40  & 100
& 100 & 100
& 100 & 100 \\

\bottomrule
\end{tabular}%
}
\caption{
Completion and group-wise detailed classification results.
Acc. denotes accuracy; NT denotes the proportion of non-\textsc{True}
predictions in the completion probe, and NC denotes the proportion of
predictions not assigned to class C in each group.
All values are percentages.
}
\label{tab:completion-group-results}
\end{table}

\begin{table}[t]
\centering

\resizebox{\columnwidth}{!}{
\begin{tabular}{lccccc}
\toprule
\textbf{Model}
& \textbf{Int.}
& \textbf{Occ.}
& \textbf{Intr.}
& \textbf{NC}
& \textbf{Comp.} \\
\midrule

Llama
& 100 & 0 & 0 & 0 & 100 \\

Qwen
& 100 & 20 & 45 & 0 & 100 \\

GLM
& 100 & 15 & 0 & 0 & 100 \\

DeepSeek
& 60 & 10 & 60 & 0 & 100 \\

\bottomrule
\end{tabular}
}
\caption {Accuracy on the occurrence experiment. Int., Occ., Intr., NC, and
Comp. denote intention, occurrence, interruption, non-completion, and
completion.
}
\label{tab:occurrence-experiment}
\end{table}

\subsection{What Underlies the Calibration Crisis?}
\begin{table}[t]
\centering

\footnotesize
\renewcommand{\arraystretch}{1.08}

\resizebox{\columnwidth}{!}{%
\begin{tabular}{lcccc}
\toprule
\textbf{Step}
& \textbf{Zero-shot}
& \textbf{DAP}
& \textbf{CoT}
& \textbf{Counterfactual} \\
\midrule

Predicate ID
& $\times$
& $\times$
& $\times$
& $\times$ \\

Aspect class.
& $\times$
& $\times$
& Unclear
& $\times$ \\

Process check
& $\times$
& \checkmark
& $\times$
& $\times$ \\

Endpoint ID
& $\times$
& $\times$
& Misleading
& $\times$ \\

Culmination check
& $\times$
& \checkmark
& \checkmark
& $\times$ \\

Non-culm. check
& $\times$
& $\times$
& $\times$
& \checkmark \\

\rowcolor{gray!8}
\textbf{Semantic rule}
& $\times$
& $\times$
& $\times$
& $\times$ \\

NLI alignment
& \checkmark
& \checkmark
& \checkmark
& \checkmark \\

\bottomrule
\end{tabular}%
}
\caption{
Coverage of reasoning steps under different prompting methods.
}
\label{tab:prompt-step-coverage}
\end{table}

Prior work reports a \textbf{{Calibration Crisis}}: prompting interventions such as DAP, chain-of-thought, and counterfactual prompting can reduce prediction errors for telic predicates while inducing overcorrection on atelic predicates, suggesting that models fail to adapt their reasoning to predicate aspect~\citep{imperfective}. However, this trade-off does not necessarily imply a corresponding change in the model's underlying event-semantic reasoning.
As summarized in Table~\ref{tab:prompt-step-coverage}, the existing prompts only partially specify the reasoning required by the task. In particular, none explicitly covers the complete chain from predicate identification and aspectual classification to semantic rule application. 
Most importantly, none explicitly specifies how the intermediate semantic judgments determine the final NLI relation. Detailed prompt analyses are provided in Appendix~\ref{sec:prompt_design}.
We therefore distinguish \textbf{Decision Shift}, in which prompting changes how available semantic evidence is mapped to output labels, from {semantic discrimination}, in which the model becomes better at making the underlying semantic distinctions.
We conduct a simple set of experiments using revised DAP and chain-of-thought prompts. We use DAPCoT as a procedural-control condition: it explicitly supplies the decomposition and decision rule needed to execute the task, allowing us to test whether errors persist once the required procedure is specified. These prompts contain all the steps and remain relatively lightweight, leaving substantial room for further optimization. The results are shown in Table~\ref{tab:main_results}. Improvements under this condition show that model performance can increase when the intended reasoning procedure is made explicit.
We further ablate three components of the linguistic guidance: \texttt{DC-n} removes the distinction between event occurrence and completion, \texttt{DC-d} removes the definitions of the relevant aspectual classes, and \texttt{DC-p} removes the mapping from these semantic properties to simple-past truth conditions. 
Overall, the ablations show that performance depends on guidance beyond culmination detection alone, particularly on correctly connecting predicate aspect and event status to the final inference relation.

\begin{table}[t]
\centering

\scriptsize
\setlength{\tabcolsep}{3.0pt}
\renewcommand{\arraystretch}{1.02}

\resizebox{\columnwidth}{!}{%
\begin{tabular}{@{}llcccccccc@{}}
\toprule
\multirow{2}{*}{\textbf{Model}}
& \multirow{2}{*}{\textbf{Method}}
& \multicolumn{4}{c}{\textbf{Orig.}}
& \multicolumn{4}{c}{\textbf{Min.}} \\
\cmidrule(lr){3-6}
\cmidrule(lr){7-10}
&
& \textbf{A} & \textbf{B} & \textbf{C} & \textbf{D}
& \textbf{A} & \textbf{B} & \textbf{C} & \textbf{D} \\
\midrule

\multirow{7}{*}{Llama}
& \texttt{ZS}
& 44 & 93 & 7 & \textbf{99}
& \cellcolor{gray!15}66
& \cellcolor{gray!15}94
& \cellcolor{gray!15}0
& \cellcolor{gray!15}\textbf{100} \\

& \texttt{DAP}
& \cellcolor{gray!15}54
& \cellcolor{gray!15}\textbf{96}
& \cellcolor{gray!15}74
& \cellcolor{gray!15}82
& 60 & \textbf{96} & 46 & 50 \\
\groupline{7-10}

& \texttt{CoT}
& \textbf{94} & 34 & 56 & 59
& 36 & 4 & \textbf{66} & 42 \\

\cmidrule(lr){2-10}

& \texttt{DC}
& 78 & 73 & \textbf{76} & 66
& 60 & 36 & 52 & 46 \\
\groupline{3-6}

& \texttt{DC-n}
& 83 & 64 & 63 & 64
& 72 & 44 & 60 & 44 \\

& \texttt{DC-d}
& 85 & 51 & 69 & 47
& \textbf{78} & 30 & 56 & 28 \\

& \texttt{DC-p}
& 77 & 18 & 40 & 32
& 54 & 24 & 26 & 58 \\

\specialrule{\lightrulewidth}{3pt}{3pt}

\multirow{7}{*}{Qwen}
& \texttt{ZS}
& 42 & \textbf{98} & 46 & \textbf{97}
& 54 & \textbf{94} & 30 & \textbf{100} \\

& \texttt{DAP}
& 73 & 97 & 91 & 79
& \textbf{74} & 84 & 78 & 52 \\
\groupline{7-10}

& \texttt{CoT}
& 84 & 78 & \textbf{97} & 38
& 66 & 70 & 84 & 16 \\

\cmidrule(lr){2-10}

& \texttt{DC}
& \cellcolor{gray!15}74
& \cellcolor{gray!15}97
& \cellcolor{gray!15}93
& \cellcolor{gray!15}81
& 28 & 88 & 90 & 66 \\
\groupline{3-6}

& \texttt{DC-n}
& 63 & 95 & 84 & 77
& \cellcolor{gray!15}52
& \cellcolor{gray!15}84
& \cellcolor{gray!15}86
& \cellcolor{gray!15}80 \\
\groupline{7-10}

& \texttt{DC-d}
& \textbf{93} & 80 & \textbf{97} & 45
& 50 & 70 & \textbf{98} & 62 \\

& \texttt{DC-p}
& 83 & 80 & 89 & 35
& 46 & 70 & 76 & 52 \\

\specialrule{\lightrulewidth}{3pt}{3pt}

\multirow{7}{*}{GLM}
& \texttt{ZS}
& 27 & 99 & 2 & \textbf{100}
& 36 & \textbf{100} & 0 & \textbf{100} \\

& \texttt{DAP}
& 30 & \textbf{100} & 24 & \textbf{100}
& 46 & \textbf{100} & 2 & 98 \\

& \texttt{CoT}
& 87 & 69 & \textbf{100} & 11
& 54 & 24 & \textbf{100} & 2 \\

\cmidrule(lr){2-10}

& \texttt{DC}
& \cellcolor{gray!15}\textbf{100}
& \cellcolor{gray!15}85
& \cellcolor{gray!15}94
& \cellcolor{gray!15}76
& \cellcolor{gray!15}86
& \cellcolor{gray!15}56
& \cellcolor{gray!15}92
& \cellcolor{gray!15}42 \\

& \texttt{DC-n}
& 98 & 87 & 81 & 82
& \textbf{96} & 54 & 60 & 48 \\
\groupline{3-6}
\groupline{7-10}

& \texttt{DC-d}
& 85 & 91 & 96 & 81
& 82 & 58 & 94 & 40 \\
\groupline{3-6}

& \texttt{DC-p}
& 77 & 76 & 51 & 59
& 56 & 48 & 8 & 64 \\

\specialrule{\lightrulewidth}{3pt}{3pt}

\multirow{7}{*}{DeepSeek}
& \texttt{R}
& 35 & \textbf{91} & 73 & 38
& 56 & \textbf{88} & 34 & 50 \\
\groupline{7-10}

& \texttt{DAP+R}
& 76 & 50 & 59 & 9
& 62 & 38 & 50 & 8 \\

& \texttt{Explanation+R}
& \textbf{88} & 31 & \textbf{93} & 9
& 62 & 12 & \textbf{78} & 2 \\

\cmidrule(lr){2-10}

& \texttt{DC}
& \cellcolor{gray!15}62
& \cellcolor{gray!15}79
& \cellcolor{gray!15}32
& \cellcolor{gray!15}79
& \textbf{82} & 60 & 24 & 74 \\

& \texttt{DC-n}
& 56 & 81 & 31 & \textbf{81}
& \cellcolor{gray!15}66
& \cellcolor{gray!15}70
& \cellcolor{gray!15}18
& \cellcolor{gray!15}\textbf{88} \\

& \texttt{DC-d}
& 53 & 75 & 29 & 72
& 68 & 60 & 24 & 68 \\

& \texttt{DC-p}
& 67 & 50 & 44 & 36
& 60 & 40 & 26 & 36 \\
\groupline{3-6}

\bottomrule
\end{tabular}%
}
\caption{
Results (\%) on the two benchmarks.
For Group A, Orig. reports the joint rate of \textsc{False} and
\textsc{Unknown} predictions, whereas Min. reports the \textsc{False} rate.
ZS, R, and DC denote zero-shot, reasoning, and DAPCoT, respectively.
Within each model and dataset, the best result for each group is shown in
\textbf{bold}; the method with the highest macro-average across
Groups A--D is shaded in gray; and the method with the highest worst-group
accuracy is marked by a line.
}
\label{tab:main_results}

\end{table}


\section{Beyond Culmination: Diagnosing Event-Semantic Reasoning}


\subsection{Intermediate-Output Evaluation}
\label{sec:intermediate_outputs}

To identify where model errors arise, we require models to explicitly output the intermediate variables needed for the final NLI decision. These include
(1) the predicted aspectual class; (2) whether the event process occurred; (3) whether culmination is entailed; (4) NLI relation; (5) the final label; (6) Code mapping with intermediate results.
All intermediate variables and the final label are elicited within a single structured output. This setting tests whether the model can execute the complete reasoning process while maintaining consistency between its intermediate judgments and final prediction.
We report both the accuracy of each intermediate variable and the consistency between the intermediate outputs and the final label. The results are in Table~\ref{tab:intermediate_compact}. The results point to two prominent sources of difficulty: aspectual classification and the downstream conversion of semantic judgments into final NLI decisions. The experiment also provides additional evidence for Sufficiency  Error, as models have higher accuracy on the completion question but lower accuracy on the final label.
Intermediate probes do not provide direct or definitive access to the model's internal reasoning, and their outputs may be influenced by prompt wording, answer format, or local task demands. We therefore treat them as diagnostic clues that help analyze possible failure modes, rather than as conclusive evidence about the model's underlying representations or reasoning process. We examine these failure modes with more evidence in the following sections.

\begin{table}[t]
\centering

\scriptsize
\setlength{\tabcolsep}{2.6pt}
\renewcommand{\arraystretch}{1.08}

\begin{tabular*}{\columnwidth}{
@{\extracolsep{\fill}}llccc|ccc@{}
}
\toprule
\textbf{Data}
& \textbf{Model}
& \multicolumn{3}{c|}{\textbf{Semantic}}
& \multicolumn{3}{c}{\textbf{Relation / Mapping}} \\
\cmidrule(lr){3-5}
\cmidrule(lr){6-8}
&
& \textbf{Tel.}
& \textbf{Proc.}
& \textbf{Comp.}
& \textbf{Rel.}
& \textbf{Final}
& \textbf{Code} \\
\midrule

\multirow{4}{*}{\textbf{Orig.}}
& Llama
& 63
& \textbf{98}
& 86
& \underline{48}
& \underline{48}
& 61 \\

& GLM
& \underline{56}
& \textbf{100}
& 99
& 64
& 64
& \underline{56} \\

& Qwen
& 81
& \textbf{100}
& 93
& 55
& \underline{54}
& 81 \\

& DeepSeek
& 67
& 79
& \textbf{97}
& \underline{44}
& \underline{44}
& 60 \\

\midrule

\multirow{4}{*}{\textbf{Min.}}
& Llama
& 61
& \textbf{96}
& 86
& 54
& \underline{49}
& 61 \\

& GLM
& 70
& \textbf{100}
& 99
& \underline{66}
& \underline{66}
& 69 \\

& Qwen
& 87
& \textbf{100}
& 92
& 54
& \underline{53}
& 88 \\

& DeepSeek
& 77
& 82
& \textbf{96}
& \underline{52}
& \underline{52}
& 65 \\

\bottomrule
\end{tabular*}
\caption{
Accuracy (\%) of intermediate semantic decisions and final predictions.
\textbf{Tel.}: telicity; \textbf{Proc.}: whether the event process occurred;
\textbf{Comp.}: whether completion is entailed; \textbf{Rel.}: predicted NLI relation;
\textbf{Final}: model's final label; \textbf{Code}: label deterministically mapped from the intermediate judgments.
Within each row, the highest value is shown in \textbf{bold} and the lowest
value is \underline{underlined}.
}
\label{tab:intermediate_compact}

\end{table}

\subsection{Aspectual Classification}
\begin{table}[t]
\centering

\small
\setlength{\tabcolsep}{4.2pt}
\renewcommand{\arraystretch}{1.03}

\begin{tabular*}{\columnwidth}{
@{\extracolsep{\fill}}llccc@{}
}
\toprule
\textbf{Model} & \textbf{Setting}
& \textbf{Orig.}
& \textbf{VSS}
& \textbf{Min.} \\
\midrule

\multirow{4}{*}{Llama}
& Zero-shot
& \textbf{74.5}
& \textbf{73.0}
& 58.0 \\

& DAP
& \underline{58.5}
& \underline{57.8}
& \underline{50.5} \\

& CoT
& 62.3
& 64.5
& 54.0 \\

& Counterf.
& 72.5
& 70.3
& \textbf{62.5} \\

\midrule

\multirow{4}{*}{Qwen}
& Zero-shot
& 83.5
& 85.8
& \textbf{74.5} \\

& DAP
& \textbf{90.0}
& \textbf{92.0}
& 71.0 \\

& CoT
& 83.3
& 83.3
& 61.5 \\

& Counterf.
& \underline{70.0}
& \underline{70.3}
& \underline{52.0} \\

\midrule

\multirow{4}{*}{GLM}
& Zero-shot
& 64.3
& 68.8
& 58.0 \\

& DAP
& \textbf{90.8}
& \textbf{94.0}
& \textbf{82.5} \\

& CoT
& 89.5
& 91.0
& 81.0 \\

& Counterf.
& \underline{50.3}
& \underline{50.0}
& \underline{51.0} \\

\midrule

\multirow{4}{*}{DeepSeek}
& Reasoning
& \underline{39.8}
& \underline{41.0}
& \underline{35.5} \\

& DAP+R
& 56.8
& 59.8
& 56.0 \\

& CoT+R
& \textbf{79.0}
& \textbf{80.5}
& 63.0 \\

& Counterf.+R
& 66.8
& 72.8
& \textbf{64.5} \\

\bottomrule
\end{tabular*}
\caption{
Macro-average accuracy (\%) for aspectual classification for each prompting setting.
Counterf. denotes counterfactual prompting.
Within each model and dataset, the highest performance is shown in
\textbf{bold}, and the lowest performance is \underline{underlined}.
}
\label{tab:c}
\end{table}

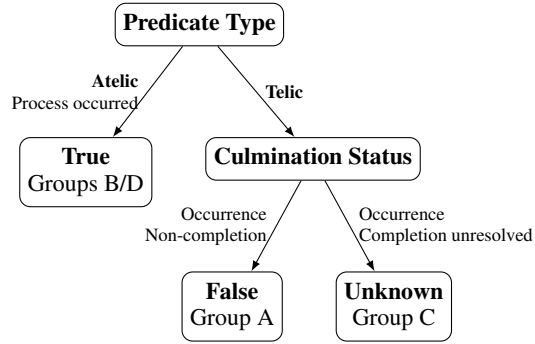
\begin{figure}[t]
\centering

\begin{forest}
for tree={
    draw,
    rounded corners,
    align=center,
    font=\footnotesize,
    edge={-latex},
    l sep=12mm,
    s sep=7mm,
    inner sep=3pt
}
[\textbf{Predicate Type}
    [\textbf{True}\\Groups B/D,
        edge label={
            node[midway,left,align=right,font=\scriptsize]
            {\textbf{Atelic}\\Process occurred}
        }
    ]
    [\textbf{Culmination Status},
        edge label={
            node[midway,right,font=\scriptsize]
            {\textbf{Telic}}
        }
        [\textbf{False}\\Group A,
            edge label={
                node[midway,left,align=right,font=\scriptsize]
                {Occurrence\\Non-completion}
            }
        ]
        [\textbf{Unknown}\\Group C,
            edge label={
                node[midway,right,align=left,font=\scriptsize]
                {Occurrence\\Completion unresolved}
            }
        ]
    ]
]
\end{forest}

\caption{Decision process for assigning labels to Groups A--D.}
\label{fig:group-decision-tree}
\end{figure}

Although aspectual classification was treated as a preliminary step in prior work~\citep{imperfective}, our experiments show that it is far from trivial for 7B--9B-scale language models. Reliably distinguishing between the two predicate classes is a necessary prerequisite for resolving the Imperfective Paradox.
As illustrated in Figure~\ref{fig:group-decision-tree}, the final three-way NLI decision begins with a binary classification of the target predicate. The two predicate classes trigger different reasoning paths and ultimately correspond to different output-label mappings. Even if all subsequent decisions are error-free, the final prediction will remain incorrect whenever the predicate is misclassified.
We evaluate aspectual classification using the original prompts from the previous work. The results show that performance is generally limited across 7B--9B models in Table \ref{tab:c}. Furthermore, we observe the same decision shift phenomenon as in the full benchmark: Some prompt modifications primarily shift the decision boundary between the two predicate classes, while yielding only modest improvements in the models' ability to distinguish telic and atelic predicates themselves. 

\subsection{Oracle-Guided Evaluation}
\begin{table}[t]
\centering

\footnotesize
\setlength{\tabcolsep}{2.5pt}
\renewcommand{\arraystretch}{1.15}

\resizebox{\columnwidth}{!}{%
\begin{tabular}{l*{7}{c}}
\toprule
\multirow{2}{*}{\textbf{Rep.}}
& \multicolumn{2}{c}{\textbf{Llama}}
& \multicolumn{2}{c}{\textbf{Qwen}}
& \multicolumn{2}{c}{\textbf{GLM}}
& \multicolumn{1}{c}{\textbf{DeepSeek}} \\
\cmidrule(lr){2-3}
\cmidrule(lr){4-5}
\cmidrule(lr){6-7}
\cmidrule(lr){8-8}

& \textbf{NR} & \textbf{R}
& \textbf{NR} & \textbf{R}
& \textbf{NR} & \textbf{R}
& \textbf{R} \\
\midrule

S$\rightarrow$S
& 59
& 98
& 98
& \textbf{100}
& \underline{50}
& \textbf{100}
& \textbf{100} \\

O$\rightarrow$S
& 43
& 88
& \textbf{100}
& \textbf{100}
& \underline{25}
& 63
& \textbf{100} \\

S$\rightarrow$O
& \underline{50}
& 98
& 75
& 56
& \underline{50}
& \textbf{100}
& 96 \\

O$\rightarrow$O
& 71
& \textbf{100}
& 75
& \textbf{100}
& \underline{50}
& 53
& \textbf{100} \\

\bottomrule
\end{tabular}%
}
\caption{
Accuracy under different representation settings.
Within each representation setting, the highest accuracy is shown in \textbf{bold}, and the lowest is
\underline{underlined}.
}
\label{tab:representation_reasoning}
\end{table}

We further conduct oracle-guided experiments to examine how final decisions change when semantic uncertainty is reduced, and the intended decision procedure is explicitly specified. To test whether models can reliably use the same underlying information across different representations, we provide the gold intermediate states together with the deterministic procedure that maps them to the final label. We then vary how both the intermediate states and final answers are expressed: either semantically (S; e.g., \textit{completed}/\textit{not completed}) or with opaque labels (O; e.g., \textit{A}/\textit{B}).
We evaluate four representation settings with and without explicit reasoning. The two terms indicate whether intermediate states and final answers use meaningful semantic labels (e.g., \textit{occurred}, \textit{completed}) or arbitrary symbols. As shown in Table~\ref{tab:representation_reasoning}, without reasoning, performance varies substantially across answer formats, suggesting sensitivity to label surface forms. Explicit reasoning generally improves performance, but the gains remain dependent on how intermediate states and outputs are represented.
These results indicate that benchmark errors do not arise solely from failures to represent the imperfective distinction. Under the controlled conditions in which the relevant semantic states and decision rules are provided, model predictions remain sensitive to how these states and output labels are represented. 


\subsection{Evaluation of Larger Models}










\begin{table}[t]
\centering

\scriptsize
\setlength{\tabcolsep}{2.5pt}
\renewcommand{\arraystretch}{1.05}

\resizebox{\columnwidth}{!}{%
\begin{tabular}{lllccccc}
\toprule
\textbf{Model}
& \textbf{Dataset}
& \textbf{Setting}
& \textbf{A}
& \textbf{B}
& \textbf{C}
& \textbf{D}
& \textbf{Avg.} \\
\midrule

\multirow{6}{*}{Qwen-72B}
& \multirow{3}{*}{Orig.}
& NR
& 91 & 91 & 77 & 99 & 89.5 \\
&
& R
& 82 & 95 & 97 & 78 & 88.0 \\
&
& 1R
& 86 & 94 & 98 & 90 & \textbf{92.0} \\

\cmidrule(lr){2-8}

& \multirow{3}{*}{Min.}
& NR
& 86 & 78 & 38 & 100 & 75.5 \\
&
& R
& 82 & 82 & 80 & 90 & 83.5 \\
&
& 1R
& 86 & 88 & 88 & 90 & \textbf{88.0} \\

\midrule

\multirow{6}{*}{GPT-5.4}
& \multirow{3}{*}{Orig.}
& NR
& 78 & 91 & 86 & 93 & 87.0 \\
&
& R
& 72 & 95 & 77 & 99 & 85.8 \\
&
& 1R
& 81 & 96 & 76 & 99 & \textbf{88.0} \\

\cmidrule(lr){2-8}

& \multirow{3}{*}{Min.}
& NR
& 86 & 86 & 48 & 92 & 78.0 \\
&
& R
& 86 & 98 & 64 & 96 & 86.0 \\
&
& 1R
& 92 & 94 & 88 & 98 & \textbf{93.0} \\

\bottomrule
\end{tabular}
}
\caption{
Results (\%) on two benchmarks.
For Group A, Orig. reports the joint rate of \textsc{False} and
\textsc{Unknown} predictions, whereas Min. reports the \textsc{False} rate.
NR, R, and 1R denote no reasoning, reasoning, and one-shot reasoning,
respectively. Avg. is the mean across Groups A--D.
}
\label{tab:large_models_compact}
\end{table}

We provide the results of an open-source LLM (Qwen2.5-72B) and a closed-source LLM (GPT-5.4) in Table \ref{tab:large_models_compact}. We find that the average performance in the four groups is still around 90 in the benchmark. However, in Group C, the reasoning traces show that the models do not judge the event to be completed, but the models still choose True in some cases.
Further inspection of the generated reasoning traces reveals that most errors do not arise from failures in the later event-semantic reasoning steps. Once a predicate has been classified, both models generally reason correctly about process occurrence, culmination, and the corresponding NLI relation. Their predictions instead diverge primarily at the aspectual classification stage.
This pattern partly reflects the context sensitivity of aspectual classification. 

\section{Conclusion}
\label{sec:conclusion}

We reexamine the existing imperfective paradox benchmark construction, analysis, and conclusions. We identify three benchmark biases: aspectual reduction, semantic bias, and relation bias, and construct lexically matched minimal pairs to better control annotation quality and lexical variation. By reformulating event-semantic NLI as a multi-step reasoning problem, we show that models often do not affirm culmination but nevertheless accept the corresponding telic simple-past statement. Our prompting and ablation results further distinguish genuine improvements in semantic reasoning from {Decision Shift} among output labels. Finally, intermediate-output and oracle-guided evaluations reveal sensitivity in aspectual classification and label representations for final label prediction. 

\section*{Limitations}

\label{sec:limitations}

Our study has several limitations.

\paragraph{Aspectual Ambiguity and Evaluation.}
Aspectual interpretation can be inherently ambiguous, as
the same predicate may admit different telic or atelic
readings depending on its linguistic and discourse context.
Although our binary telic--atelic classification provides
a useful operationalization for controlled evaluation,
it may not fully capture such interpretive variation.
Future work could explore graded judgments of aspectual
interpretation and entailment acceptability, as well as
distributions of human judgments, to complement categorical
accuracy. Another direction is to examine the trade-off
between three-way NLI classification (True, False, and
Unknown) and binary entailment/non-entailment evaluation at the same time,
as the latter may reduce sensitivity to disputed labels
while obscuring the distinction between contradiction
and uncertainty. Please refer to Appendix~\ref{app:a1} for more discussion.

\paragraph{The Imperfective Paradox as a Semantic Benchmark for LLMs.}
The imperfective paradox captures an important but specific
phenomenon in event semantics, raising the question of
whether benchmarks built around this phenomenon can
adequately evaluate LLMs' broader semantic competence instead of handling of this particular contrast or following the artificial rules.
Moreover, our study adopts a strict semantic entailment
standard, whereas ordinary language users may draw plausible
pragmatic inferences that are not logically entailed.
Comparing formal semantic judgments with naturally elicited
human interpretations could help distinguish genuine
reasoning failures from differences in pragmatic
understanding.


A further complication is that aspectual interpretation is compositional and interacts with world knowledge and this may lead to ambiguity in some cases. For LLMs, frequencies and associations in pretraining text may serve as proxies for such knowledge in some cases: responses to progressive--simple-past pairs may reflect the attestation of these constructions across events and participants. Such influences are not inherently biases, since human interpretation also draws on experience and expectations about how events typically unfold. The use of ``semantic judgments'' describes the content of elicited responses. The claim concerns the diagnostic insufficiency of such disagreement, while any systematic model tendency needs further examination.
Please refer to
Appendix~\ref{app:a2} for more discussion.

\paragraph{World Knowledge and Complex Context.}
Our evaluation primarily uses short, controlled contexts and limits assumptions about event outcomes that are not established by the premise. This restriction may ignore that aspectual interpretation is independent of world knowledge. Aspectual properties are compositionally determined, while contextual and event knowledge can shape the available interpretations. Naturally occurring language additionally involves broader discourse and interacting contextual cues that our examples only partially capture. Future work should examine how the observed response patterns change in richer, knowledge-intensive contexts, including cases where background expectations support or conflict with explicitly stated outcomes.

\paragraph{Cross-Linguistic Generalization.}
Our study focuses on English, while aspectual distinctions
and culmination inferences can vary across languages.
Some languages permit non-culminating interpretations
of event descriptions that would normally imply completion
in English. Extending the analysis to languages with
different aspectual systems could help determine whether
the observed failure patterns reflect general semantic
reasoning difficulties or language-specific conventions.

\paragraph{Controlled Minimal Pairs.}
Our controlled examples are designed to address aspectual
reduction by considering predicate-level aspect, isolate
particular semantic contrasts, and reduce lexical and
contextual variation. While this improves diagnostic
clarity, it may not capture the full range of interpretations
available in naturally occurring language, especially
for predicates with less clearly determined aspectual
interpretations.

\paragraph{Human Annotation Protocol.}
Our human annotation is conducted under a single
background-informed setting: annotators are introduced to
the relevant linguistic background and annotation criteria
before making their judgments. We do not compare this
protocol against other plausible annotation settings,
including (1) zero-shot annotation without guidance,
(2) annotation with only task instructions but no explicit
linguistic rules, and (3) re-annotation in which annotators
first provide an initial judgment and then reconsider
the same examples after being introduced to the relevant
linguistic distinctions and rules. Such comparisons would
help determine to what extent the resulting labels depend
on the annotation protocol itself.

\paragraph{Simplified Linguistic Framework.}
The linguistic background and rules introduced in our study
provide a simplified account of a substantially more complex
theoretical literature. Aspectual interpretation and
culmination judgments can depend on lexical semantics,
argument structure, discourse context, and the adopted
linguistic analysis. Our operationalization should therefore
be viewed as a controlled approximation for analysis.

\paragraph{Intermediate Judgments as Behavioral Evidence.}
Intermediate outputs provide only behavioral evidence about
possible failure modes and may be sensitive to prompt wording
and answer format. They should therefore not be interpreted
as direct evidence of the models' internal semantic
representations or reasoning processes. Further interpretability research is needed to investigate
how these semantic distinctions are internally represented
and used in model reasoning.

\section*{Acknowledgement}
We thank Katarzyna Pruś for her comments on the earlier draft of this paper.

\bibliography{custom}

\appendix

\label{sec:appendix}


\section{Further Discussion}
\label{telic}

\subsection{Discussion: Everyday English and Languages with Weak Culmination Inferences}
\label{ch}



Many languages permit non-culminating interpretations of accomplishment predicates, including Mandarin Chinese, Hindi, Tamil, and several Salish languages, although the availability and grammatical source of such interpretations vary across languages. In these languages, a perfective or otherwise bounded event description does not always guarantee that the event reached its inherent endpoint, as shown in Box~\ref{box:cross-linguistic-culmination}~\citep{ArunachalamKothari2011,noncul}\footnote{The linguistic background and underlying explanation are more complex than presented here. We refer readers to the cited work for a more detailed discussion.}. Although canonical English simple-past accomplishments generally entail culmination, telicity is not morphologically marked on the verb itself and must be computed from the complete predicate and its linguistic and discourse context
~\citep{filip2008,bott2024processing}. 
Consequently, the distinction between event occurrence and event completion may not always be represented consistently in the multilingual and informal text on which LLMs are trained. Differences in how languages encode culmination, together with multilingual parameter sharing, may make this distinction more difficult for models to learn reliably
~\citep{xue2021mt5massivelymultilingualpretrained,dodge-etal-2021-documenting,wang-etal-2020-negative,pfeiffer2022liftingcursemultilingualitypretraining}.

\begin{tcolorbox}[
title=\textbf{Cross-Linguistic Contrast in Culmination},
colback=gray!5,
colframe=gray!70,
boxrule=0.5pt,
arc=2mm,
left=2mm,
right=2mm,
top=1mm,
bottom=1mm
]
\label{box:cross-linguistic-culmination}

\small

\textbf{Mandarin Chinese: Acceptable}

\begin{quote}
\textit{Zhāngsān shā-le Lǐsì liǎng cì, kěshì Lǐsì méi sǐ.}

Literal: ``Zhangsan killed Lisi two times, but Lisi did not die.''
\end{quote}


\medskip

\textbf{English: Unacceptable}

\begin{quote}
James killed Hannibal twice, but Hannibal did not die.
\end{quote}

\end{tcolorbox}

\subsection{Beyond Categorical Aspectual and NLI Evaluation}
\label{app:a1}

Our evaluation adopts a binary distinction between telic
and atelic predicates and primarily measures performance
through categorical accuracy. While this design is
appropriate for isolating controlled semantic contrasts,
aspectual interpretation itself can be ambiguous.
Certain predicates may admit multiple plausible readings
depending on argument structure, contextual completion
standards, and discourse information. Consequently,
assigning a single aspectual class may not always capture
the range of interpretations available to human speakers.
Importantly, such variation does not necessarily imply
that telicity is intrinsically continuous; it may instead
reflect competing discrete interpretations.

An interesting direction for future work is to investigate
whether graded judgments of aspectual interpretation and
entailment acceptability can complement categorical accuracy.
Evaluating models against distributions of human judgments,
rather than a single gold label, could further help
distinguish genuine semantic errors from differences among
plausible linguistic interpretations.

A related question concerns the granularity of NLI labels.
Reducing the three-way True/False/Unknown classification
to a binary entailment/non-entailment distinction may
provide a useful complementary evaluation, particularly
when the distinction between contradiction and uncertainty
is difficult to establish. However, this reduction also
removes information about whether a hypothesis is
contradicted or merely unsupported by the premise.
Comparing binary and three-way evaluations could therefore
reveal how sensitive model assessments are to label
granularity and help clarify which semantic distinctions
models reliably capture.
\subsection{The Imperfective Paradox as a Semantic Benchmark for LLMs}
\label{app:a2}

The imperfective paradox captures an important but specific
phenomenon in event semantics, raising a broader question
about the suitability of using benchmarks built around
this phenomenon to evaluate LLMs' semantic competence.
While such benchmarks provide a useful diagnostic of
particular semantic distinctions, it remains unclear
to what extent their results reflect broader semantic
understanding. In particular, one important issue
concerns the distinction between formal semantic entailment
and pragmatic understanding. Under strict NLI semantics,
an interrupted event does not necessarily imply
non-completion, whereas ordinary language users may
pragmatically infer that the intended outcome was not
achieved. Such inferences can be reasonable in everyday
communication even when they are not logically entailed
by the premise. This raises the question of whether models
should be evaluated solely on formal semantic correctness
or also on their alignment with ordinary human
interpretations. Future work could investigate this
distinction through complementary evaluations based on
strict logical entailment and naturally elicited human
judgments, examining when models follow formal semantic
rules and when they rely on plausible pragmatic inferences.
Such an approach could help distinguish failures of logical
reasoning from differences in pragmatic interpretation
without treating either as universally preferable.

World knowledge also shapes the interpretation of event descriptions. Aspectual interpretation is compositional and interacts with knowledge about typical event developments. For LLMs, frequencies and distributional associations in pretraining text may serve as proxies for such knowledge, influencing judgments about progressive and simple-past descriptions. Human interpretations likewise draw on expectations about how events usually unfold. This suggests examining how humans and models combine compositional cues with event expectations, particularly when contextual evidence reinforces or overrides a typical outcome. Such comparisons could clarify how these sources of information contribute to judgments of logical entailment, likely outcomes, and intended meaning.
\subsection{Telicity}
\label{te}
Canonical contrasts such as \textit{build a house} versus \textit{run in the
park} are relatively clear in English. The former describes an event with an
inherent culmination, whereas the latter
describes an activity without a linguistically encoded endpoint. However, the
boundary between telic and atelic predicates is not always categorical, and
aspectual class should not be treated as a fixed property of an isolated verb.
Rather, it is determined compositionally by the complete predicate, including
the verb, its arguments, quantization, path expressions, and result structure
\citep{filip1999aspect,TelicityandTerminativity}. For example, \textit{write a
ten-page report} is naturally telic, whereas \textit{write in a notebook} is
atelic; similarly, \textit{walk to the station} introduces an endpoint, while
\textit{walk in the station} does not.

The effect of a nominal argument on telicity
depends on the semantic relation between that argument and the event, rather
than on the argument's surface form alone. With incremental-theme predicates,
the parts of the relevant participant correspond to successive parts of the
event. Thus, \textit{eat a sandwich} is normally telic because the gradual
consumption of the bounded sandwich measures the progress of the eating event,
whereas \textit{eat blueberries} is normally atelic because the bare-plural
object does not provide a fixed upper bound. However, a quantized object does
not automatically produce a telic predicate. Predicates such as \textit{watch
the documentary} and \textit{like the documentary} remain atelic because the
documentary does not measure or delimit the temporal development of the
watching or liking eventuality \citep{filip1999aspect,hand,HUUMO}. This distinction shows
that benchmark annotation cannot infer telicity merely from the presence of a
definite, singular, or quantified object.

Cases in which the same verb participates in predicates with different aspectual interpretations involve what is often called aspect shift or aspectual coercion. For example, an
inherently activity-like verb may occur in the telic predicates
\textit{run to the store}, \textit{laugh oneself silly}, or
\textit{wheeze upward to the fourteenth floor}. In English, this shift between
atelic and telic interpretations is generally not morphologically marked on
the verb; instead, it is induced by result phrases, bounded paths, arguments,
and linguistic or extra-linguistic context \citep{filip1999aspect,HUUMO}. A
verb-level classifier may therefore assign the same aspectual class to
\textit{run in the park} and \textit{run to the store}, although only the
latter predicate linguistically introduces a destination that delimits the
event.


Maximalization analysis provides a more general account of these
patterns. On this view, telic predicates denote events that are maximal
relative to a contextually and linguistically supplied ordering scale. Events do not culminate independently of a particular
description: the same drinking event, for example, may be maximal relative to
\textit{drink a glass of wine} but non-maximal relative to
\textit{drink a bottle of wine}. 
Some predicates also permit more than one aspectual construal. Expressions such
as \textit{clean the garage}, \textit{wash the car}, or \textit{pick
strawberries} may be interpreted as bounded tasks in one context but as
open-ended activities in another. 
Degree achievements and incremental-theme
predicates introduce an additional complication because their culmination
conditions may depend on the relevant scale, the degree to which an object is
affected, and a contextually determined standard of maximality. Thus, the presence of an object or an
intended real-world goal does not by itself guarantee that the linguistic
predicate is telic.
In particular, an action may be performed in pursuit of an
external goal without encoding that goal as part of its truth conditions \citep{filip1999aspect,hand,HUUMO}.


These distinctions create unavoidable uncertainty for benchmark annotation.
When a predicate has multiple linguistically plausible construals, disagreement
with a single gold label need not constitute a genuine semantic error. 
\section{Detailed Discussion about three Validity Concerns}
\subsection{Aspectual Reduction}
\label{sec:predicate_aspect}

The first concern is in the benchmark's representation of lexical aspect. 
Aspectual class is not generally an invariant property of an isolated verb. Rather, it is compositionally determined by the full predicate, including the verb, its arguments, quantization properties, path or goal expressions, and other syntactic constituents. The same lexical verb can consequently participate in predicates with different aspectual properties. For example, \textit{run} in \textit{run in the park} ordinarily describes an atelic activity, whereas \textit{run home} or \textit{run a mile} introduces a bounded path or extent and can receive a telic interpretation. Similarly, \textit{pick apples} is typically atelic, while \textit{pick an apple} is associated with an incremental theme and a natural endpoint~\citep{filip1999aspect}. Therefore, the following examples in the benchmark may permit different label interpretations.



\begin{quote}
\label{ex:b007}

\noindent
\textbf{Example (a): \texttt{A\_098}}\\
\textbf{Premise:} We were picking strawberries, but a bear chased us off.\\
\textbf{Hypothesis:} We picked strawberries.\\
\textbf{Original label:} \textsc{False}

\vspace{0.5em}

\noindent
\textbf{Example (b): \texttt{B\_007}}\\
\textbf{Premise:} The choir was singing a hymn, but the conductor sneezed.\\
\textbf{Hypothesis:} The choir sang a hymn.\\
\textbf{Original label:} \textsc{True}

\end{quote}
\subsection{Semantic Bias}
\label{sec:completion_semantics}

The second concern involves both the semantic criterion used to determine whether a telic event counts as completed and the fact that some predicate expressions admit both atelic and telic uses. The benchmark generally assumes that completion requires the full realization of a canonical endpoint.
This assumption is too strong for many natural-language predicates. Completion judgments can depend on the lexical semantics of the predicate, the structure of its incremental theme, the degree to which the relevant object has been affected, and the contextual standard adopted by conversational participants. For predicates such as \textit{eat an apple}, the progress of the event can be measured against the affected portions of the object. Nevertheless, the context may influence what degree of realization is treated as sufficient, and ordinary speakers may accept that someone ate an apple even when a contextually negligible piece remains~\citep{filip1999aspect,filip2008}.
A further complication is that some predicate expressions admit both atelic and telic use. The benchmark often resolves such ambiguity in favor of a single strict telic reading. A model that adopts an activity-like or less strongly resultative construal may therefore receive a different label, even when its interpretation is linguistically plausible~\citep{filip1999aspect,filip2008,article}. Therefore, the following examples in the benchmark may receive different labels under linguistically plausible interpretations. 


\begin{quote}
\label{ex:c018-c098}

\noindent
\textbf{Example (c): \texttt{C\_018}}\\
\textbf{Premise:} He was cleaning the garage.\\
\textbf{Hypothesis:} He cleaned the garage.\\
\textbf{Original label:} \textsc{Unknown}

\vspace{0.5em}

\noindent
\textbf{Example (d): \texttt{C\_022}}\\
\textbf{Premise:} He was washing the car.\\
\textbf{Hypothesis:} He washed the car.\\
\textbf{Original label:} \textsc{Unknown}

\end{quote}
This does not mean that all partial events satisfy their corresponding completed predicates. A builder who constructs only the foundation has ordinarily not \textit{built a house}. 
Task completion should distinguish substantial non-completion from contextually negligible deviations, rather than relying on a uniform all-or-nothing rule.
If multiple interpretations are permitted, two distinct sources of unknown should be distinguished. See Appendix C for this distinction and our annotation procedure.



\subsection{Relation Bias}
\label{sec:interruption_nli}

The third concern arises in the conversion of event descriptions into NLI labels. The benchmark frequently treats evidence that an event was interrupted as evidence that the corresponding completed-event statement is false. This inference is not valid without additional information about the eventual outcome of the event.
Consider a premise stating that a storm interrupted the construction of a gazebo. The interruption establishes that construction did not proceed continuously during the described interval. It does not establish that the builder never resumed construction or that the gazebo was not eventually completed. The completed-event hypothesis is therefore not entailed, but neither is it contradicted. Under standard three-way NLI semantics, the appropriate relation is \textsc{Neutral}.
Furthermore, an interruption may occur only after the endpoint relevant to the hypothesis has already been reached. Indeed, such descriptions do not merely remain compatible with event completion; the explicit reference to a completed result often provides direct evidence that the relevant endpoint was reached before the subsequent interruption or destruction occurred. For example, a premise may state that ``the student finished writing the essay, but accidentally deleted the file before submitting it.'' The deletion interrupts the broader process of completing and submitting the assignment, but it does not negate the fact that the essay-writing event had already culminated. Two examples from the dataset are shown as follows~\citep{prus-etal-2025-world}.

\begin{quote}
\label{ex:a094-a081}

\noindent
\textbf{Example (a): \texttt{A\_094}}\\
\textbf{Premise:} He was shaping the surfboard, but cut too deep.\\
\textbf{Hypothesis:} He shaped the surfboard.\\
\textbf{Original label:} \textsc{False}

\vspace{0.5em}

\noindent
\textbf{Example (b): \texttt{A\_081}}\\
\textbf{Premise:} She was cutting the fabric for the dress, but realized she
cut it wrong.\\
\textbf{Hypothesis:} She cut the fabric for the dress.\\
\textbf{Original label:} \textsc{False}

\end{quote}


\section{Data}

We conduct our experiments on three versions of ImperfectiveNLI: the
\emph{Original}, \emph{VSS}, and \emph{Lexically matched Minimal-Pair} datasets. The
Original dataset contains 400 examples, evenly distributed across four
groups: interrupted accomplishments (Group A), interrupted activities
(Group B), ambiguous accomplishments (Group C), and ambiguous activities
(Group D), with 100 examples per group. The VSS dataset removes 97
examples identified as containing annotation concerns, resulting in 303 examples: 49, 98, 58, and 98 examples in Groups
A--D, respectively. See Appendix~\ref{human} for additional discussion of annotation concerns. The Minimal-Pair dataset contains 200 newly constructed
examples, with 50 examples per group. Its examples are organized as
controlled minimal pairs that vary only the event-semantic information
relevant to the target inference, thereby reducing the influence of lexical
and contextual differences. We use the Original dataset and the Lexically Matched Minimal-Pair dataset as the primary controlled diagnostic. The VSS is reported only as an auxiliary sensitivity analysis to illustrate model behavior after removing examples with identified annotation concerns because its construction involves selective filtering, which may introduce bias and can not serve as an independent source of evidence for our main conclusions. Unless otherwise stated, we evaluate
each model on all available examples in each dataset.

The VSS dataset is obtained by excluding the 97 examples listed. The complete
list of excluded example IDs is provided in the box below.
\begin{tcolorbox}[
title=\textbf{Examples 1},
colback=gray!5,
colframe=gray!70,
boxrule=0.5pt,
arc=2mm,
left=2mm,
right=2mm,
top=1mm,
bottom=1mm,
breakable
]
\small

\noindent
{\ttfamily
A\_003, A\_007, A\_009, A\_013, A\_015, A\_018, A\_021, A\_022,
A\_023, A\_024, A\_025, A\_026, A\_027, A\_030, A\_031, A\_032,
A\_033, A\_034, A\_035, A\_039, A\_043, A\_047, A\_050, A\_054,
A\_055, A\_058, A\_061, A\_064, A\_067, A\_069, A\_070, A\_073,
A\_074, A\_075, A\_076, A\_077, A\_078, A\_079, A\_081, A\_082,
A\_083, A\_084, A\_085, A\_088, A\_089, A\_090, A\_094, A\_096,
A\_097, A\_098, A\_100, B\_007, B\_079, C\_008, C\_009, C\_015,
C\_018, C\_021, C\_022, C\_023, C\_024, C\_025, C\_026, C\_027,
C\_029, C\_033, C\_034, C\_035, C\_036, C\_039, C\_043, C\_050,
C\_055, C\_061, C\_073, C\_074, C\_075, C\_076, C\_077, C\_078,
C\_079, C\_081, C\_082, C\_083, C\_084, C\_085, C\_088, C\_089,
C\_090, C\_091, C\_094, C\_096, C\_097, C\_098, C\_100, D\_007,
D\_079.
}

\end{tcolorbox}

The examples without explicit evidence of non-completion are listed below. This structural criterion is stricter than the human-annotation criterion. Absence of explicit non-completion does not by itself imply that annotators will reject the original label.
\begin{tcolorbox}[
title=\textbf{Examples 2},
colback=gray!5,
colframe=gray!70,
boxrule=0.5pt,
arc=2mm,
left=2mm,
right=2mm,
top=1mm,
bottom=1mm,
breakable
]
\small

\noindent
{\ttfamily
A\_002, A\_003, A\_007, A\_009, A\_011, A\_013, A\_014, A\_015,
A\_016, A\_018, A\_019, A\_021, A\_022, A\_023, A\_024, A\_025,
A\_026, A\_027, A\_030, A\_031, A\_032, A\_033, A\_034, A\_035,
A\_036, A\_037, A\_038, A\_039, A\_042, A\_043, A\_045, A\_046,
A\_047, A\_050, A\_054, A\_055, A\_058, A\_059, A\_061, A\_062,
A\_063, A\_064, A\_065, A\_067, A\_068, A\_069, A\_070, A\_071,
A\_072, A\_073, A\_074, A\_075, A\_076, A\_077, A\_078, A\_079,
A\_080, A\_081, A\_082, A\_083, A\_084, A\_085, A\_086, A\_087,
A\_088, A\_089, A\_090, A\_091, A\_092, A\_093, A\_094, A\_096,
A\_097, A\_098, A\_099, A\_100
}

\end{tcolorbox}



\paragraph{Lexically matched Minimal-Pair}
To obtain a clean diagnostic setting, our minimal pairs deliberately use predicates with relatively explicit path, quantity, or result structure, thereby reducing the context-dependent completion ambiguity. This design is intended as an evaluation control rather than a general theory of culmination semantics; examples are additionally constructed to make the relevant endpoint explicit and to minimize residual partial-completion readings. When outcome information is required, completion or non-completion is stated explicitly in the premise rather than inferred from interruption or contextual plausibility.

In detail, our lexically matched minimal pairs use explicit path,
quantity, or result expressions to produce clearer telic--atelic contrasts
while holding the lexical verb and general scenario constant. To minimize
residual interpretive ambiguity, we adopt a deliberately conservative
criterion: an example is retained only when the intended aspectual contrast
and the relevant culmination status are explicitly supported by the complete
predicate and context, rather than inferred from typicality or world
knowledge. We also explicitly point out incompleteness in the context for certain groups. This design does not assume that every aspectual distinction is
categorically fixed; instead, it focuses the controlled evaluation on cases
for which the intended interpretation is least controversial. We further
validate the resulting examples through human annotation to verify that the
intended predicate classifications and NLI relations are reliably supported. Two annotators agree on all the labels with independent labeling. 

\subsection{Implications for Annotation Validity}

These findings suggest an {annotation-validity gap} rather than uniform
label noise. If the annotation procedure itself were generally unreliable, we
would expect comparable instability across the four conditions. Instead,
Groups B and D show near-complete agreement with the intended labels, whereas
the disagreement is concentrated in Groups A and C.

This asymmetry matters for interpreting model errors. A model--gold
disagreement on an activity example can usually be interpreted against a
relatively stable human judgment. The same inference is less straightforward
for a substantial subset of accomplishment examples, where multiple human
annotators accept an alternative relation or fail to reproduce the original
label under majority voting. Consequently, benchmark accuracy on these items
can conflate genuine model errors with differences in linguistically plausible
predicate or culmination interpretations.

For this reason, the human annotations should not be treated as establishing a new
single gold label for every disputed example. Instead, we use them to identify
where categorical evaluation is well supported and where the benchmark
contains substantial interpretive uncertainty. We
need to distinguish strict evaluation under the original labels from more
permissive analyses that explicitly preserve alternative interpretations when
they are relevant to the semantic question under study.
\section{Human Annotation}
\label{human}
The analysis raises a concern about annotation validity.
This question is particularly important for accomplishment predicates, where
the interpretation of culmination may depend on the complete predicate,
contextual standards of completion, and what is inferred from an interruption.
We therefore conduct a human annotation study over the full benchmark. Our goal is not to replace the original
labels with a new categorical annotation scheme, but to measure how strongly
the original single-label judgments are supported and whether alternative
interpretations are distributed uniformly across the four benchmark
conditions.

\subsection{Annotation Protocol}

Three native English-speaking annotators independently annotated all 400 premise--hypothesis pairs,
with 100 examples in each of Groups A--D. The examples are shuffled randomly. The original benchmark assigns
\textsc{False} to interrupted accomplishments (A), \textsc{True} to
interrupted activities (B), \textsc{Unknown} to outcome-underspecified
accomplishments (C), and \textsc{True} to outcome-underspecified activities
(D).

In the annotation, we first introduce the linguistic background of the problem to every annotator. Please see our discussion about other annotation settings in Limitations.
Rather than forcing annotators to select exactly one label, we allowed them to
mark every NLI relation they considered linguistically acceptable:
\textsc{True}, \textsc{False}, and/or \textsc{Unknown}. This design is intended
to preserve cases in which more than one interpretation is judged possible,
rather than converting interpretive variation into artificial disagreement.
When an annotator selected multiple acceptable labels, we additionally asked
for a \emph{tendency} judgment indicating which interpretation they preferred.
This allows us to analyze the annotations at two levels. First, the
\emph{acceptable-label} analysis preserves all interpretations considered
acceptable by an annotator. Second, the \emph{preferred-label} analysis collapses
each annotation to a single label using the annotator's stated tendency. All
multi-label annotations in our data have a unique tendency, so this procedure
produces one preferred label for every annotator and every example.
We report two complementary measures. An example is counted as
\emph{admitting an alternative interpretation} when at least two of the three
annotators consider at least one label other than the original gold label
acceptable. We additionally perform majority voting over the annotators'
preferred labels. These two analyses distinguish the existence of a plausible
alternative interpretation from the stronger case in which annotators'
preferred judgments fail to reproduce the original label.

Note on the interpretation of \textsc{Unknown}.
In our annotation scheme, \textsc{Unknown} represents uncertainty about the outcome of an event under a fixed telic interpretation: the predicate is taken to encode an inherent endpoint, but the premise does not establish whether that endpoint was reached. We distinguish this from uncertainty about the predicate's aspectual interpretation itself. If, for instance, an atelic construal supports \textsc{True} while a telic construal supports \textsc{False}, the annotation should preserve both labels rather than collapse the two readings into \textsc{Unknown}. This distinction allows the multi-label annotation to separate \emph{culmination uncertainty} from \emph{aspectual ambiguity}.

Our use of background-informed annotation follows established practice in expert linguistic annotation, where annotators are often familiarized with the relevant theoretical distinctions and annotation criteria before making judgments on semantically complex phenomena. The purpose of this procedure is to ensure that annotators distinguish the linguistic notions relevant to the task rather than relying on potentially inconsistent interpretations of the terminology. Providing such background may not be methodologically neutral: introducing linguistic distinctions may influence how annotators construe ambiguous examples. We view background-informed annotation as a deliberate trade-off between theoretically controlled judgments and possible protocol-induced priming, rather than as a procedure that eliminates annotation bias. Importantly, our protocol does not force annotators toward a single theory-consistent label. Annotators may select all NLI relations they consider linguistically acceptable and separately indicate their preferred, or tendency, interpretation. This multi-label-plus-tendency design is deliberately conservative: the acceptable-label component preserves plausible alternative interpretations, while the tendency judgment still requires each annotator to commit to the interpretation they consider most natural. Consequently, the protocol does not mechanically convert every theoretically available alternative into disagreement with the original label, and the preferred judgments remain directly comparable to a conventional single-label annotation setting. The combination therefore allows us to measure interpretive variation without giving up the information that would have been obtained had annotators been required to choose only one label.
\subsection{Alternative Interpretations Are Concentrated in Accomplishments}

Table~\ref{tab:human_annotation} summarizes the main results. Across the full
benchmark, at least two of three annotators considered an alternative to the
original gold label acceptable for 70 of 400 examples (\textbf{17.5\%}).
Importantly, however, this uncertainty is not distributed uniformly across the
benchmark.

In Group A, at least two annotators supported an alternative interpretation for
\textbf{38\%} of the examples, and the corresponding proportion is
\textbf{29\%} in Group C. By contrast, alternative interpretations are rare in
the two activity conditions: only \textbf{2\%} of Group B and \textbf{1\%} of
Group D receive alternative-label support from at least two annotators.
Pooling the conditions, \textbf{33.5\%} of accomplishment examples
(A/C; 67 of 200) admit an alternative interpretation under this criterion,
compared with only \textbf{1.5\%} of activity examples
(B/D; 3 of 200).

\begin{table*}[t]
\centering
\small
\begin{tabular}{lcccc}
\toprule
 & A & B & C & D \\
\midrule
At least two annotators support
an alternative label
& 38\% & 2\% & 29\% & 1\% \\

Majority vote reproduces
original gold
& 71\% & 98\% & 71\% & 99\% \\
\bottomrule
\end{tabular}
\caption{
Human annotation results by benchmark group.
The first row reports the proportion of examples for which at least two of
three annotators considered a label other than the original gold label
acceptable. The second row reports the proportion for which majority voting
over annotators' preferred interpretations (Only one for each annotator) reproduces the original gold
label.
}
\label{tab:human_annotation}
\end{table*}

The same asymmetry remains when we require annotators to commit to a preferred
interpretation. Majority voting reproduces the original gold label for only
\textbf{71\%} of both Groups A and C, compared with \textbf{98\%} and
\textbf{99\%} for Groups B and D, respectively. Thus, the result cannot be
explained solely by our permissive multi-label annotation interface: even after each annotator's response is reduced to a single preferred judgment, the original labels
remain substantially less stable in the accomplishment conditions.



This distinction is important. Our human results do not establish that every
disputed original label is incorrect. In many cases, the original label remains
one linguistically acceptable interpretation. The stronger conclusion
supported by the data is that the assumption of a uniquely determined
categorical gold label is considerably less secure in the accomplishment
conditions that carry the benchmark's central semantic contrast.

\section{Additional Results}

\subsection{Completion and Occurrence Analysis}
This subsection provides the full results for the completion and
occurrence-to-completion analyses discussed in the main text.
The completion and group-wise results are reported in
Table~\ref{tab:completion-group-results1}, and the full occurrence experiment
is reported in Table~\ref{tab:occurrence-experiment1}.
\begin{table}[t]
\centering

\resizebox{\columnwidth}{!}{
\begin{tabular}{llcc|cc|cc|cc|cc}
\toprule
\multirow{3}{*}{\textbf{Model}}
& \multirow{3}{*}{\textbf{Prompt}}
& \multicolumn{2}{c|}{\multirow{2}{*}{\textbf{Completion}}}
& \multicolumn{8}{c}{\textbf{Group-wise Results}} \\
\cmidrule(lr){5-12}
&
&
&
& \multicolumn{2}{c|}{\textbf{A}}
& \multicolumn{2}{c|}{\textbf{B}}
& \multicolumn{2}{c|}{\textbf{C}}
& \multicolumn{2}{c}{\textbf{D}} \\
\cmidrule(lr){3-4}
\cmidrule(lr){5-6}
\cmidrule(lr){7-8}
\cmidrule(lr){9-10}
\cmidrule(lr){11-12}
&
& \textbf{Acc.} & \textbf{NT}
& \textbf{Acc.} & \textbf{NC Rate}
& \textbf{Acc.} & \textbf{NC Rate}
& \textbf{Acc.} & \textbf{NC Rate}
& \textbf{Acc.} & \textbf{NC Rate} \\
\midrule

\multirow{2}{*}{Llama}
& No reasoning
& 0 & 100
& 0 & 100
& 0 & 100
& 90 & 90
& 100 & 100 \\

& Reasoning
& 85 & 95
& 75 & 100
& 60 & 95
& 85 & 85
& 85 & 85 \\

\midrule

\multirow{2}{*}{Qwen}
& No reasoning
& 45 & 100
& 70 & 85
& 0 & 40
& 65 & 65
& 10 & 10 \\

& Reasoning
& 100 & 100
& 85 & 100
& 45 & 100
& 100 & 100
& 100 & 100 \\

\midrule

\multirow{2}{*}{GLM}
& No reasoning
& 0 & 100
& 0 & 70
& 0 & 100
& 100 & 100
& 100 & 100 \\

& Reasoning
& 80 & 100
& 55 & 100
& 15 & 100
& 100 & 100
& 100 & 100 \\

\midrule

\multirow{2}{*}{DeepSeek}
& No reasoning
& 55 & 100
& 65 & 100
& 30 & 90
& 100 & 100
& 100 & 100 \\

& Reasoning
& 90 & 100
& 80 & 100
& 40 & 100
& 100 & 100
& 100 & 100 \\

\bottomrule
\end{tabular}
}
\caption{Results on the completion and group-wise classification experiments.
For the completion experiment, Acc. denotes accuracy and NT denotes the
proportion of predictions that are not classified as \textsc{True}.
For each group, NC Rate denotes the proportion of predictions not assigned to
class C. All values are percentages.}
\label{tab:completion-group-results1}

\end{table}

\begin{table}[t]
\centering

\resizebox{\columnwidth}{!}{
\begin{tabular}{lccccc}
\toprule
\textbf{Model}
& \textbf{Int.}
& \textbf{Occ.}
& \textbf{Intr.}
& \textbf{NC}
& \textbf{Comp.} \\
\midrule

Llama
& 100 & 0 & 0 & 0 & 100 \\

Qwen
& 100 & 20 & 45 & 0 & 100 \\

GLM
& 100 & 15 & 0 & 0 & 100 \\

DeepSeek
& 60 & 10 & 60 & 0 & 100 \\

\bottomrule
\end{tabular}
}
\caption{Results on the occurrence experiment. Int., Occ., Intr., NC, and
Comp. denote intention, occurrence, interruption, non-completion, and
completion, respectively. The reported result is accuracy. The ground-truth
label is \textsc{True} only in the Completion set.}
\label{tab:occurrence-experiment1}

\end{table}

\subsection{Prompting and Ablation Results}
This subsection reports the full group-wise results for the prompting
experiments and DAPCoT ablations discussed in the main text.
Results across the Original benchmark, VSS, and Minimal Pairs are provided.
\begin{table*}[t]
\centering

\resizebox{\textwidth}{!}{
\begin{tabular}{llcccccc|cccccc|cccccc}
\toprule
\multirow{2}{*}{\textbf{Model}}
& \multirow{2}{*}{\textbf{Method}}
& \multicolumn{6}{c|}{\textbf{Original}}
& \multicolumn{6}{c|}{\textbf{VSS}}
& \multicolumn{6}{c}{\textbf{Minimal Pairs}} \\
\cmidrule(lr){3-8}
\cmidrule(lr){9-14}
\cmidrule(lr){15-20}
&
& \textbf{A} & \textbf{B} & \textbf{C} & \textbf{D}
& \textbf{A-F} & \textbf{A-U}
& \textbf{A} & \textbf{B} & \textbf{C} & \textbf{D}
& \textbf{A-F} & \textbf{A-U}
& \textbf{A} & \textbf{B} & \textbf{C} & \textbf{D}
& \textbf{A-F} & \textbf{A-U} \\
\midrule

\multirow{8}{*}{Llama-3.1-8B-Instruct}
& \texttt{Zero-shot}
& 44 & 93 & 7 & 99 & 17 & 27
& 59 & 92 & 7 & 99 & 29 & 31
& 78 & 94 & 0 & 100 & 66 & 12 \\
& \texttt{DAP}
& 54 & 96 & 74 & 82 & 15 & 39
& 76 & 95 & 83 & 85 & 27 & 49
& 74 & 96 & 46 & 50 & 60 & 14 \\
& \texttt{CoT}
& 94 & 34 & 56 & 59 & 6 & 88
& 100 & 37 & 67 & 58 & 10 & 90
& 100 & 4 & 66 & 42 & 36 & 64 \\
& \texttt{Counterfactual}
& 99 & 1 & 95 & 1 & 44 & 55
& 98 & 0 & 95 & 0 & 63 & 35
& 100 & 0 & 90 & 0 & 68 & 32 \\
\cmidrule(lr){2-20}
& \texttt{DAPCoT}
& 78 & 73 & 76 & 66 & 38 & 40
& 86 & 73 & 74 & 72 & 49 & 37
& 88 & 36 & 52 & 46 & 60 & 28 \\
& \texttt{DAPCoT-n}
& 83 & 64 & 63 & 64 & 47 & 36
& 88 & 63 & 72 & 74 & 57 & 31
& 94 & 44 & 60 & 44 & 72 & 22 \\
& \texttt{DAPCoT-d}
& 85 & 51 & 69 & 47 & 53 & 32
& 94 & 53 & 55 & 38 & 65 & 29
& 86 & 30 & 56 & 28 & 78 & 8 \\
& \texttt{DAPCoT-p}
& 77 & 18 & 40 & 32 & 24 & 53
& 82 & 13 & 50 & 37 & 35 & 47
& 88 & 24 & 26 & 58 & 54 & 34 \\
\midrule

\multirow{8}{*}{Qwen2.5-7B-Instruct}
& \texttt{Zero-shot}
& 42 & 98 & 46 & 97 & 21 & 21
& 53 & 98 & 55 & 97 & 35 & 18
& 62 & 94 & 30 & 100 & 54 & 8 \\
& \texttt{DAP}
& 73 & 97 & 91 & 79 & 35 & 38
& 88 & 97 & 93 & 79 & 53 & 35
& 80 & 84 & 78 & 52 & 74 & 6 \\
& \texttt{CoT}
& 84 & 78 & 97 & 38 & 29 & 55
& 92 & 78 & 97 & 39 & 45 & 47
& 88 & 70 & 84 & 16 & 66 & 22 \\
& \texttt{Counterfactual}
& 86 & 45 & 100 & 1 & 8 & 78
& 90 & 47 & 100 & 2 & 14 & 76
& 90 & 68 & 98 & 6 & 32 & 58 \\
\cmidrule(lr){2-20}
& \texttt{DAPCoT}
& 74 & 97 & 93 & 81 & 18 & 56
& 86 & 98 & 97 & 83 & 22 & 63
& 86 & 88 & 90 & 66 & 28 & 58 \\
& \texttt{DAPCoT-n}
& 63 & 95 & 84 & 77 & 40 & 23
& 84 & 95 & 88 & 77 & 59 & 24
& 74 & 84 & 86 & 80 & 52 & 22 \\
& \texttt{DAPCoT-d}
& 93 & 80 & 97 & 45 & 15 & 78
& 94 & 82 & 98 & 49 & 27 & 67
& 82 & 70 & 98 & 62 & 50 & 32 \\
& \texttt{DAPCoT-p}
& 83 & 80 & 89 & 35 & 11 & 72
& 82 & 80 & 86 & 33 & 20 & 61
& 74 & 70 & 76 & 52 & 46 & 28 \\
\midrule

\multirow{8}{*}{GLM-4-9B}
& \texttt{Zero-shot}
& 27 & 99 & 2 & 100 & 20 & 7
& 27 & 97 & 3 & 98 & 24 & 2
& 40 & 100 & 0 & 100 & 36 & 4 \\
& \texttt{DAP}
& 30 & 100 & 24 & 100 & 26 & 4
& 39 & 100 & 21 & 100 & 35 & 4
& 46 & 100 & 2 & 98 & 46 & 0 \\
& \texttt{CoT}
& 87 & 69 & 100 & 11 & 31 & 56
& 90 & 60 & 100 & 9 & 43 & 47
& 90 & 24 & 100 & 2 & 54 & 36 \\
& \texttt{Counterfactual}
& 95 & 6 & 100 & 0 & 16 & 79
& 88 & 9 & 100 & 1 & 25 & 63
& 98 & 16 & 100 & 0 & 26 & 72 \\
\cmidrule(lr){2-20}
& \texttt{DAPCoT}
& 100 & 85 & 94 & 76 & 66 & 34
& 100 & 87 & 91 & 72 & 82 & 18
& 94 & 56 & 92 & 42 & 86 & 8 \\
& \texttt{DAPCoT-n}
& 98 & 87 & 81 & 82 & 89 & 9
& 100 & 90 & 86 & 87 & 94 & 6
& 96 & 54 & 60 & 48 & 96 & 0 \\
& \texttt{DAPCoT-d}
& 85 & 91 & 96 & 81 & 60 & 25
& 90 & 90 & 98 & 81 & 65 & 24
& 96 & 58 & 94 & 40 & 82 & 14 \\
& \texttt{DAPCoT-p}
& 77 & 76 & 51 & 59 & 30 & 47
& 88 & 79 & 53 & 49 & 39 & 49
& 76 & 48 & 8 & 64 & 56 & 20 \\
\midrule

\multirow{8}{*}{DeepSeek-R1-7B}
& \texttt{Reasoning}
& 35 & 91 & 73 & 38 & 8 & 27
& 55 & 90 & 74 & 39 & 16 & 39
& 76 & 88 & 34 & 50 & 56 & 20 \\
& \texttt{DAP+R}
& 76 & 50 & 59 & 9 & 30 & 46
& 84 & 49 & 66 & 11 & 51 & 33
& 92 & 38 & 50 & 8 & 62 & 30 \\
& \texttt{Explanation+R}
& 88 & 31 & 93 & 9 & 32 & 56
& 92 & 33 & 90 & 12 & 39 & 53
& 98 & 12 & 78 & 2 & 62 & 36 \\
& \texttt{Counterfactual+R}
& 69 & 57 & 14 & 34 & 54 & 15
& 82 & 58 & 9 & 33 & 67 & 14
& 88 & 56 & 2 & 32 & 70 & 18 \\
\cmidrule(lr){2-20}
& \texttt{DAPCoT}
& 62 & 79 & 32 & 79 & 36 & 26
& 71 & 85 & 38 & 76 & 39 & 33
& 90 & 60 & 24 & 74 & 82 & 8 \\
& \texttt{DAPCoT-n}
& 56 & 81 & 31 & 81 & 36 & 20
& 63 & 85 & 41 & 82 & 41 & 22
& 78 & 70 & 18 & 88 & 66 & 12 \\
& \texttt{DAPCoT-d}
& 53 & 75 & 29 & 72 & 28 & 25
& 67 & 81 & 28 & 73 & 37 & 31
& 90 & 60 & 24 & 68 & 68 & 22 \\
& \texttt{DAPCoT-p}
& 67 & 50 & 44 & 36 & 18 & 49
& 49 & 56 & 38 & 30 & 16 & 33
& 72 & 40 & 26 & 36 & 60 & 12 \\

\bottomrule
\end{tabular}
}
\caption{Performance across the original, VSS, and minimal pairs.
 A-F and A-U denote the percentages of
\textsc{False} and \textsc{Unknown} predictions for Group A, respectively,
while A denotes their sum, i.e., the proportion of non-\textsc{True}
predictions, computed before rounding. All values are percentages rounded to integers.}
\label{tab:main_results1}
\end{table*}

We conduct a simple set of experiments using revised DAP and chain-of-thought prompts. We use DAPCoT as a procedural-control condition: it explicitly supplies the decomposition and decision rule needed to execute the task, allowing us to test whether errors persist once the required procedure is specified. These prompts contain all the steps and remain relatively lightweight, leaving substantial room for further optimization, such as providing clearer definitions of the two predicate types, adding more detailed steps, and incorporating few-shot demonstrations, particularly examples with complete reasoning traces. The results are shown in Table~\ref{tab:main_results}. These gains are not obviously a uniform shift toward a particular label.
We ablate three pieces of linguistic guidance from the revised prompt. \texttt{DAPCoT-n} removes the explicit statement that the progressive form establishes event occurrence but does not by itself entail completion. \texttt{DAPCoT-d} removes the definitions of the two relevant event classes, namely atelic activities and telic accomplishments. \texttt{DAPCoT-p} further removes the distinction between their simple-past truth conditions: occurrence is sufficient for an atelic predicate, whereas culmination is normally required for a telic predicate.
The results show that these components play complementary roles. Removing the progressive-form guidance mainly shifts predictions between \textsc{False} and \textsc{Unknown} in Group A, suggesting that it helps distinguish event occurrence from completion. Removing the event-class definitions causes broader, model-dependent degradation, especially for groups requiring recognition of atelic predicates. The largest decline occurs when the mapping from semantic facts to simple-past truth conditions is removed, indicating that identifying occurrence or culmination alone is insufficient; models must also understand how these properties determine the final inference relation.
DeepSeek behaves differently from the other models. In the zero-shot setting, it sometimes treats interruption as evidence of completion, while under DAPCoT it often fails to follow the provided reasoning procedure and applies inappropriate semantic rules. This suggests that reasoning-oriented models may require more explicit and carefully designed instructions for this task.






\subsection{Intermediate-output Evaluation}
This subsection provides the full intermediate-output evaluation underlying
the stage-wise diagnostic analysis in the main text.
Results for all intermediate semantic decisions and final predictions across
the three datasets are reported in Table~\ref{tab:intermediate_results1}.

\begin{table*}[t]
\centering

\setlength{\tabcolsep}{2.2pt}
\resizebox{\textwidth}{!}{
\begin{tabular}{
l
cccccccc|
cccccccc|
cccccccc
}
\toprule
\multirow{2}{*}{\textbf{Model}}
& \multicolumn{8}{c|}{\textbf{Original}}
& \multicolumn{8}{c|}{\textbf{VSS}}
& \multicolumn{8}{c}{\textbf{Minimal Pairs}} \\
\cmidrule(lr){2-9}
\cmidrule(lr){10-17}
\cmidrule(lr){18-25}

& \textbf{All}
& \textbf{Tel.}
& \textbf{Proc.}
& \textbf{Comp.}
& \textbf{Int.}
& \textbf{Rel.}
& \textbf{Final}
& \textbf{Code}

& \textbf{All}
& \textbf{Tel.}
& \textbf{Proc.}
& \textbf{Comp.}
& \textbf{Int.}
& \textbf{Rel.}
& \textbf{Final}
& \textbf{Code}

& \textbf{All}
& \textbf{Tel.}
& \textbf{Proc.}
& \textbf{Comp.}
& \textbf{Int.}
& \textbf{Rel.}
& \textbf{Final}
& \textbf{Code} \\
\midrule

\textbf{Llama-3.1-8B}
& 50 & 63 & 98 & 86 & 83 & 48 & 48 & 61
& 52 & 64 & 100 & 85 & 82 & 48 & 45 & 57
& 46 & 61 & 96 & 86 & 81 & 54 & 49 & 61 \\

\textbf{GLM-4-9B}
& 48 & 56 & 100 & 99 & 88 & 64 & 64 & 56
& 56 & 72 & 100 & 99 & 84 & 62 & 62 & 68
& 59 & 70 & 100 & 99 & 86 & 66 & 66 & 69 \\

\textbf{Qwen2.5-7B}
& 69 & 81 & 100 & 93 & 91 & 55 & 54 & 81
& 69 & 86 & 99 & 99 & 84 & 50 & 50 & 84
& 73 & 87 & 100 & 92 & 91 & 54 & 53 & 88 \\

\textbf{DeepSeek-R1-7B}
& 43 & 67 & 79 & 97 & 81 & 44 & 44 & 60
& 41 & 61 & 83 & 94 & 79 & 56 & 54 & 54
& 50 & 77 & 82 & 96 & 83 & 52 & 52 & 65 \\
\bottomrule
\end{tabular}
}
\caption{Accuracy of intermediate semantic decisions and final NLI predictions
across the three datasets. All denotes the proportion
of outputs with intermediate labels all correct. Tel., Proc., Comp., Int., Rel.,
Final, and Code denote telicity, process occurrence, completion, interruption,
predicted NLI relation, model-final prediction, and code-mapped prediction,
respectively. All values are integer percentages.}
\label{tab:intermediate_results1}
\end{table*}
\subsection{Aspectual Classification}
This subsection reports the full results for the predicate-level aspectual
classification analysis discussed in the main text.
Group-wise results across prompting settings and all three datasets are
provided in Table~\ref{tab:main_results1}.

\begin{table*}[t]
\centering

\resizebox{\textwidth}{!}{
\begin{tabular}{llcccc|cccc|cccc}
\toprule
\multirow{2}{*}{\textbf{Model}}
& \multirow{2}{*}{\textbf{Method}}
& \multicolumn{4}{c|}{\textbf{Original}}
& \multicolumn{4}{c|}{\textbf{VSS}}
& \multicolumn{4}{c}{\textbf{Minimal Pairs}} \\
\cmidrule(lr){3-6}
\cmidrule(lr){7-10}
\cmidrule(lr){11-14}
&
& \textbf{A} & \textbf{B} & \textbf{C} & \textbf{D}
& \textbf{A} & \textbf{B} & \textbf{C} & \textbf{D}
& \textbf{A} & \textbf{B} & \textbf{C} & \textbf{D} \\
\midrule

\multirow{4}{*}{Llama-3.1-8B-Instruct}
& Zero-shot
& 95 & 52 & 96 & 55
& 96 & 46 & 97 & 53
& 96 & 22 & 96 & 18 \\
& DAP
& 100 & 18 & 100 & 16
& 100 & 16 & 100 & 15
& 96 & 4 & 98 & 4 \\
& CoT
& 99 & 28 & 100 & 22
& 100 & 29 & 100 & 29
& 100 & 6 & 100 & 10 \\
& Counterfactual
& 100 & 43 & 99 & 48
& 100 & 41 & 97 & 43
& 100 & 24 & 100 & 26 \\
\midrule

\multirow{4}{*}{Qwen2.5-7B-Instruct}
& Zero-shot
& 91 & 76 & 91 & 76
& 96 & 77 & 93 & 77
& 92 & 58 & 92 & 56 \\
& DAP
& 95 & 85 & 95 & 85
& 98 & 86 & 98 & 86
& 78 & 64 & 78 & 64 \\
& CoT
& 100 & 66 & 100 & 67
& 100 & 66 & 100 & 67
& 100 & 22 & 100 & 24 \\
& Counterfactual
& 100 & 40 & 100 & 40
& 100 & 40 & 100 & 41
& 100 & 4 & 100 & 4 \\
\midrule

\multirow{4}{*}{GLM-4-9B}
& Zero-shot
& 28 & 100 & 29 & 100
& 41 & 100 & 34 & 100
& 18 & 100 & 14 & 100 \\
& DAP
& 86 & 97 & 84 & 96
& 94 & 96 & 90 & 96
& 70 & 92 & 74 & 94 \\
& CoT
& 93 & 85 & 94 & 86
& 98 & 83 & 97 & 86
& 92 & 68 & 98 & 66 \\
& Counterfactual
& 1 & 100 & 0 & 100
& 0 & 100 & 0 & 100
& 0 & 100 & 4 & 100 \\
\midrule

\multirow{4}{*}{DeepSeek-R1-7B}
& Reasoning
& 6 & 74 & 8 & 71
& 6 & 70 & 12 & 76
& 10 & 68 & 4 & 60 \\
& DAP+R
& 46 & 68 & 42 & 71
& 41 & 69 & 53 & 76
& 36 & 58 & 50 & 80 \\
& CoT+R
& 87 & 74 & 81 & 74
& 88 & 76 & 84 & 74
& 86 & 38 & 88 & 40 \\
& Counterfactual+R
& 50 & 81 & 52 & 84
& 59 & 85 & 60 & 87
& 68 & 68 & 60 & 62 \\

\bottomrule
\end{tabular}
}
\caption{Classification accuracy (\%) for Groups A--D across three datasets.}
\label{tab:main_results1}
\end{table*}

\subsection{Oracle-guided Evaluation}
This subsection provides the full oracle-guided results underlying the
representation and reasoning analysis in the main text.
Performance for all semantic/opaque representation combinations, with and
without explicit reasoning, is reported in
Table~\ref{tab:prompt-reasoning-results1}.
\begin{table*}[t]
\centering

\resizebox{\textwidth}{!}{
\begin{tabular}{llccccc|ccccc}
\toprule
\multirow{2}{*}{\textbf{Model}}
& \multirow{2}{*}{\textbf{Prompt}}
& \multicolumn{5}{c|}{\textbf{No Reasoning}}
& \multicolumn{5}{c}{\textbf{Reasoning}} \\
\cmidrule(lr){3-7}
\cmidrule(lr){8-12}
&
& \textbf{A} & \textbf{B} & \textbf{C} & \textbf{D} & \textbf{Avg.}
& \textbf{A} & \textbf{B} & \textbf{C} & \textbf{D} & \textbf{Avg.} \\
\midrule

\multirow{4}{*}{Llama-3.1-8B-Instruct}
& S$\rightarrow$S
& 100 & 30 & 85 & 20 & 59
& 100 & 100 & 90 & 100 & 98 \\
& O$\rightarrow$S
& 0 & 15 & 100 & 55 & 43
& 100 & 80 & 100 & 70 & 88 \\
& S$\rightarrow$O
& 100 & 0 & 100 & 0 & 50
& 90 & 100 & 100 & 100 & 98 \\
& O$\rightarrow$O
& 100 & 50 & 100 & 35 & 71
& 100 & 100 & 100 & 100 & 100 \\

\midrule

\multirow{4}{*}{Qwen2.5-7B-Instruct}
& S$\rightarrow$S
& 100 & 100 & 90 & 100 & 98
& 100 & 100 & 100 & 100 & 100 \\
& O$\rightarrow$S
& 100 & 100 & 100 & 100 & 100
& 100 & 100 & 100 & 100 & 100 \\
& S$\rightarrow$O
& 100 & 100 & 0 & 100 & 75
& 10 & 100 & 15 & 100 & 56 \\
& O$\rightarrow$O
& 100 & 100 & 0 & 100 & 75
& 100 & 100 & 100 & 100 & 100 \\

\midrule

\multirow{4}{*}{GLM-4-9B}
& S$\rightarrow$S
& 100 & 0 & 100 & 0 & 50
& 100 & 100 & 100 & 100 & 100 \\
& O$\rightarrow$S
& 0 & 0 & 100 & 0 & 25
& 100 & 20 & 100 & 30 & 63 \\
& S$\rightarrow$O
& 100 & 0 & 100 & 0 & 50
& 100 & 100 & 100 & 100 & 100 \\
& O$\rightarrow$O
& 100 & 0 & 100 & 0 & 50
& 100 & 0 & 100 & 10 & 53 \\

\midrule

\multirow{4}{*}{DeepSeek-R1-7B}
& S$\rightarrow$S
& -- & -- & -- & -- & --
& 100 & 100 & 100 & 100 & 100 \\
& O$\rightarrow$S
& -- & -- & -- & -- & --
& 100 & 100 & 100 & 100 & 100 \\
& S$\rightarrow$O
& -- & -- & -- & -- & --
& 90 & 100 & 95 & 100 & 96\\
& O$\rightarrow$O
& -- & -- & -- & -- & --
& 100 & 100 & 100 & 100 & 100 \\
\bottomrule
\end{tabular}
}
\caption{Oracle-guided performance under different prompt and reasoning
settings. S and O denote semantic and opaque representations, respectively.}
\label{tab:prompt-reasoning-results1}

\end{table*}
\subsection{Evaluation of Larger Models}
This subsection reports the full evaluation of the larger models discussed
in the main text.
Results for Qwen2.5-72B and GPT-5.4 across the Original benchmark, VSS, and
Minimal Pairs are provided in Table~\ref{tab:large-model-results1}.
\begin{table*}[t]
\centering

\resizebox{\textwidth}{!}{
\begin{tabular}{lllcccccc}
\toprule
\textbf{Model}
& \textbf{Dataset}
& \textbf{Setting}
& \textbf{A}
& \textbf{B}
& \textbf{C}
& \textbf{D}
& \textbf{A-F}
& \textbf{A-U} \\
\midrule

\multirow{9}{*}{Qwen2.5-72B}
& \multirow{3}{*}{Original}
& \texttt{No reasoning}
& 91 & 91 & 77 & 99 & 40 & 51 \\
&
& \texttt{Reasoning}
& 82 & 95 & 97 & 78 & 29 & 53 \\
&
& \texttt{One-shot reasoning}
& 86 & 94 & 98 & 90 & 31 & 55 \\
\cmidrule(lr){2-9}

& \multirow{3}{*}{VSS}
& \texttt{No reasoning}
& 98 & 92 & 76 & 98 & 59 & 39 \\
&
& \texttt{Reasoning}
& 92 & 94 & 97 & 74 & 51 & 41 \\
&
& \texttt{One-shot reasoning}
& 94 & 95 & 100 & 88 & 47 & 47 \\
\cmidrule(lr){2-9}

& \multirow{3}{*}{Minimal pairs}
& \texttt{No reasoning}
& 94 & 78 & 38 & 100 & 86 & 8 \\
&
& \texttt{Reasoning}
& 94 & 82 & 80 & 90 & 82 & 12 \\
&
& \texttt{One-shot reasoning}
& 94 & 88 & 88 & 90 & 86 & 8 \\
\midrule

\multirow{9}{*}{GPT-5.4}
& \multirow{3}{*}{Original}
& \texttt{No reasoning}
& 78 & 91 & 86 & 93 & 25 & 53 \\
&
& \texttt{Reasoning}
& 72 & 95 & 77 & 99 & 13 & 59 \\
&
& \texttt{One-shot reasoning}
& 81 & 96 & 76 & 99 & 20 & 61 \\
\cmidrule(lr){2-9}

& \multirow{3}{*}{VSS}
& \texttt{No reasoning}
& 88 & 91 & 90 & 94 & 43 & 45 \\
&
& \texttt{Reasoning}
& 84 & 95 & 93 & 100 & 31 & 53 \\
&
& \texttt{One-shot reasoning}
& 92 & 96 & 91 & 99 & 37 & 55 \\
\cmidrule(lr){2-9}

& \multirow{3}{*}{Minimal pairs}
& \texttt{No reasoning}
& 92 & 86 & 48 & 92 & 86 & 6 \\
&
& \texttt{Reasoning}
& 90 & 98 & 64 & 96 & 86 & 4 \\
&
& \texttt{One-shot reasoning}
& 98 & 94 & 88 & 98 & 92 & 6 \\

\bottomrule
\end{tabular}
}
\caption{Performance of large language models across the original, VSS,
and lexically matched minimal-pair datasets. A-F and A-U denote the
percentages of \textsc{False} and \textsc{Unknown} predictions for Group A,
respectively, while A denotes their sum. All values are percentages rounded
to integers.}
\label{tab:large-model-results1}
\end{table*}

\subsection{Sensitivity to Prompt Wording and Repeated Runs}
\label{app:prompt_sensitivity}

Because diagnostic probes may themselves be sensitive to prompt wording, we
examine whether the main behavioral patterns depend on a particular prompt
formulation. We conduct two complementary analyses. First, we repeat the
original completion and event-status diagnostics using the same examples
and decoding configuration in the main experiments. We similarly repeat the structured
intermediate-output evaluation. Second, we construct two semantically
equivalent prompt variants that preserve the task definition and required
outputs while changing the surface formulation. For the repeated
original-prompt experiments, we report the mean across the three runs, with
run-level variance shown in brackets.

\paragraph{Completion and event-status diagnostics.}
Tables~\ref{tab:completion_sensitivity_original}--\ref{tab:completion_sensitivity_v2}
report the completion and event-status results.
Exact three-way classification accuracy can vary with prompt wording,
particularly in the no-reasoning condition. Importantly, however, changes in
exact completion accuracy often reflect differences in how models distinguish
\textsc{False} from \textsc{Unknown}, rather than a reversal toward a
completed-event interpretation. Under the original prompt, the non-\textsc{True}
(NT) rate remains at or near 100\% despite variation in exact completion
accuracy, and this pattern largely persists under the two alternative prompt
formulations.

The pattern is especially consistent when explicit reasoning is elicited.
Across the original prompt and both variants, all four models obtain a high
NT rate in the reasoning condition. For Group A, the non-completion (NC) rate
is high for every model under all three reasoning-prompt
formulations, although exact Group-A accuracy still varies. Thus, the central
observation does not require the model to produce the identical
\textsc{False}/\textsc{Unknown} distinction across prompts: models can avoid
classifying the event as completed while still making an incorrect downstream
NLI decision.
\begin{table*}[t]
\centering

\scriptsize
\setlength{\tabcolsep}{2.4pt}
\renewcommand{\arraystretch}{1.08}
\resizebox{\textwidth}{!}{%
\begin{tabular}{llcc|cc|cc|cc|cc}
\toprule
\textbf{Model} & \textbf{Prompt}
& \multicolumn{2}{c|}{\textbf{Completion}}
& \multicolumn{2}{c|}{\textbf{A}}
& \multicolumn{2}{c|}{\textbf{B}}
& \multicolumn{2}{c|}{\textbf{C}}
& \multicolumn{2}{c}{\textbf{D}} \\
\cmidrule(lr){3-4}
\cmidrule(lr){5-6}
\cmidrule(lr){7-8}
\cmidrule(lr){9-10}
\cmidrule(lr){11-12}
& & Acc. & NT & Acc. & NC & Acc. & NC & Acc. & NC & Acc. & NC \\
\midrule
Llama
& NR
& 0.00 [0.0000] & 100.00 [0.0000]
& 0.00 [0.0000] & 100.00 [0.0000]
& 0.00 [0.0000] & 100.00 [0.0000]
& 90.00 [0.0000] & 90.00 [0.0000]
& 100.00 [0.0000] & 100.00 [0.0000] \\
& R
& 86.67 [1.3889] & 96.67 [1.3889]
& 73.33 [5.5556] & 100.00 [0.0000]
& 63.33 [5.5556] & 93.33 [5.5556]
& 86.67 [5.5556] & 86.67 [5.5556]
& 86.67 [5.5556] & 86.67 [5.5556] \\
\midrule
Qwen
& NR
& 45.00 [0.0000] & 100.00 [0.0000]
& 70.00 [0.0000] & 83.33 [5.5556]
& 0.00 [0.0000] & 36.67 [22.2222]
& 63.33 [5.5556] & 63.33 [5.5556]
& 10.00 [0.0000] & 10.00 [0.0000] \\
& R
& 100.00 [0.0000] & 100.00 [0.0000]
& 83.33 [5.5556] & 100.00 [0.0000]
& 50.00 [16.6667] & 100.00 [0.0000]
& 100.00 [0.0000] & 100.00 [0.0000]
& 96.67 [5.5556] & 96.67 [5.5556] \\
\midrule
GLM
& NR
& 0.00 [0.0000] & 100.00 [0.0000]
& 0.00 [0.0000] & 73.33 [22.2222]
& 0.00 [0.0000] & 100.00 [0.0000]
& 100.00 [0.0000] & 100.00 [0.0000]
& 100.00 [0.0000] & 100.00 [0.0000] \\
& R
& 83.33 [18.0556] & 100.00 [0.0000]
& 56.67 [5.5556] & 100.00 [0.0000]
& 16.67 [5.5556] & 100.00 [0.0000]
& 100.00 [0.0000] & 100.00 [0.0000]
& 100.00 [0.0000] & 100.00 [0.0000] \\
\midrule
DeepSeek
& NR
& 58.33 [22.2222] & 100.00 [0.0000]
& 60.00 [16.6667] & 100.00 [0.0000]
& 33.33 [22.2222] & 90.00 [0.0000]
& 96.67 [5.5556] & 100.00 [0.0000]
& 100.00 [0.0000] & 100.00 [0.0000] \\
& R
& 88.33 [1.3889] & 100.00 [0.0000]
& 73.33 [38.8889] & 100.00 [0.0000]
& 36.67 [22.2222] & 100.00 [0.0000]
& 96.67 [5.5556] & 96.67 [5.5556]
& 100.00 [0.0000] & 100.00 [0.0000] \\
\bottomrule
\end{tabular}
}
\caption{
Sensitivity of the completion and event-status diagnostics under the original
prompt.
Values are means across the three runs, with run-level variance in brackets.
\textbf{Comp.} denotes the completion probe, \textbf{NT} the proportion of
non-\textsc{True} completion judgments, and \textbf{NC} the proportion of
predictions not assigned to the completed-event class in the event-status
diagnostic. All values are percentages.
}
\label{tab:completion_sensitivity_original}
\end{table*}

\begin{table*}[t]
\centering

\scriptsize
\setlength{\tabcolsep}{3.0pt}
\renewcommand{\arraystretch}{1.08}
\begin{tabular*}{\textwidth}{
@{\extracolsep{\fill}}llcc|cc|cc|cc|cc@{}
}
\toprule
\textbf{Model} & \textbf{Prompt}
& \multicolumn{2}{c|}{\textbf{Completion}}
& \multicolumn{2}{c|}{\textbf{A}}
& \multicolumn{2}{c|}{\textbf{B}}
& \multicolumn{2}{c|}{\textbf{C}}
& \multicolumn{2}{c}{\textbf{D}} \\
\cmidrule(lr){3-4}
\cmidrule(lr){5-6}
\cmidrule(lr){7-8}
\cmidrule(lr){9-10}
\cmidrule(lr){11-12}
& & Acc. & NT & Acc. & NC & Acc. & NC & Acc. & NC & Acc. & NC \\
\midrule
Llama    & NR & 0  & 100 & 0  & 100 & 0  & 100 & 100 & 100 & 100 & 100 \\
         & R  & 95 & 100 & 60 & 100 & 50 & 100 & 90  & 90  & 100 & 100 \\
\midrule
Qwen     & NR & 100 & 100 & 60 & 90  & 0  & 80  & 100 & 100 & 70  & 70 \\
         & R  & 100 & 100 & 70 & 100 & 30 & 100 & 100 & 100 & 100 & 100 \\
\midrule
GLM      & NR & 5  & 60  & 0  & 90  & 0  & 100 & 100 & 100 & 100 & 100 \\
         & R  & 90 & 100 & 60 & 100 & 30 & 100 & 100 & 100 & 100 & 100 \\
\midrule
DeepSeek & NR & 95 & 100 & 50 & 100 & 50 & 100 & 100 & 100 & 80  & 80 \\
         & R  & 100 & 100 & 70 & 100 & 40 & 100 & 100 & 100 & 100 & 100 \\
\bottomrule
\end{tabular*}
\caption{
Sensitivity of the completion and event-status diagnostics under prompt
variant~1. Abbreviations follow
Table~\ref{tab:completion_sensitivity_original}. All values are percentages.
}
\label{tab:completion_sensitivity_v1}
\end{table*}

\begin{table*}[t]
\centering

\scriptsize
\setlength{\tabcolsep}{3.0pt}
\renewcommand{\arraystretch}{1.08}
\begin{tabular*}{\textwidth}{
@{\extracolsep{\fill}}llcc|cc|cc|cc|cc@{}
}
\toprule
\textbf{Model} & \textbf{Prompt}
& \multicolumn{2}{c|}{\textbf{Completion}}
& \multicolumn{2}{c|}{\textbf{A}}
& \multicolumn{2}{c|}{\textbf{B}}
& \multicolumn{2}{c|}{\textbf{C}}
& \multicolumn{2}{c}{\textbf{D}} \\
\cmidrule(lr){3-4}
\cmidrule(lr){5-6}
\cmidrule(lr){7-8}
\cmidrule(lr){9-10}
\cmidrule(lr){11-12}
& & Acc. & NT & Acc. & NC & Acc. & NC & Acc. & NC & Acc. & NC \\
\midrule
Llama    & NR & 30  & 100 & 0  & 100 & 0  & 100 & 80  & 100 & 90  & 90 \\
         & R  & 100 & 100 & 80 & 100 & 90 & 100 & 100 & 100 & 100 & 100 \\
\midrule
Qwen     & NR & 100 & 100 & 70 & 70  & 0  & 10  & 100 & 100 & 50  & 50 \\
         & R  & 100 & 100 & 80 & 100 & 50 & 90  & 100 & 100 & 100 & 100 \\
\midrule
GLM      & NR & 100 & 100 & 0  & 30  & 0  & 80  & 100 & 100 & 90  & 90 \\
         & R  & 100 & 100 & 70 & 100 & 30 & 100 & 90  & 100 & 100 & 100 \\
\midrule
DeepSeek & NR & 60  & 100 & 60 & 100 & 30 & 100 & 100 & 100 & 80  & 80 \\
         & R  & 100 & 100 & 60 & 100 & 30 & 100 & 100 & 100 & 80  & 80 \\
\bottomrule
\end{tabular*}
\caption{
Sensitivity of the completion and event-status diagnostics under prompt
variant~2. Abbreviations follow
Table~\ref{tab:completion_sensitivity_original}. All values are percentages.
}
\label{tab:completion_sensitivity_v2}
\end{table*}

\paragraph{Intermediate-output evaluation.}
Tables~\ref{tab:intermediate_sensitivity_original}--\ref{tab:intermediate_sensitivity_v2}
show the corresponding results for the structured intermediate-output
evaluation. Although individual scores vary across prompt formulations, the
stage-wise pattern is consistent. Process-occurrence and completion judgments
are substantially more accurate than the predicted NLI relation and final
answer. Across all four models and all three prompt formulations, completion
accuracy ranges from 78.70\% to 100\%. For every model under every formulation, completion
accuracy exceeds final accuracy.

Relation accuracy also closely tracks final-answer accuracy. Across all
model--prompt combinations, the two differ by at most approximately three
percentage points. The
difficulty appears earlier in the downstream decision process, where semantic
information must be converted into the appropriate NLI relation. The
code-mapped predictions recover part of this gap for several models, although
the magnitude of the recovery varies across models and prompt formulations.

Overall, our experiemnts support the qualitative
conclusion : across repeated runs and alternative
prompt formulations, models can make relatively accurate occurrence and
culmination judgments while remaining substantially less reliable at
converting those judgments into the appropriate NLI decision. The
characteristic final error therefore should not be interpreted as direct
evidence of an erroneous culmination judgment alone.

\begin{table}[t]
\centering

\scriptsize
\setlength{\tabcolsep}{2.5pt}
\renewcommand{\arraystretch}{1.08}
\resizebox{\columnwidth}{!}{%
\begin{tabular}{lcccccc}
\toprule
\textbf{Model} & \textbf{Tel.} & \textbf{Proc.} & \textbf{Comp.}
& \textbf{Rel.} & \textbf{Final} & \textbf{Code} \\
\midrule
Llama
& 64.69 [1.3710]
& 97.83 [0.5123]
& 86.81 [0.2636]
& 47.50 [4.8230]
& 46.66 [8.3312]
& 61.60 [0.4559] \\
Qwen
& 79.18 [1.3419]
& 99.71 [0.0001]
& 93.89 [0.7087]
& 54.54 [0.1196]
& 54.90 [0.3312]
& 79.47 [1.9762] \\
GLM
& 56.50 [0.2886]
& 100.00 [0.0000]
& 99.45 [0.0513]
& 65.41 [2.3462]
& 65.41 [2.3462]
& 55.50 [0.1194] \\
DeepSeek
& 68.30 [1.1726]
& 79.88 [1.8894]
& 96.14 [0.3988]
& 46.50 [5.8547]
& 45.67 [4.2973]
& 58.91 [1.7822] \\
\bottomrule
\end{tabular}
}
\caption{
Sensitivity of the structured intermediate-output evaluation under the
original prompt. Values are means across the three runs, with run-level variance in
brackets. \textbf{Tel.}, \textbf{Proc.}, \textbf{Comp.}, \textbf{Rel.},
\textbf{Final}, and \textbf{Code} denote telicity, process occurrence,
completion, predicted NLI relation, model-final prediction, and code-mapped
prediction, respectively.
}
\label{tab:intermediate_sensitivity_original}
\end{table}

\begin{table}[t]
\centering

\scriptsize
\setlength{\tabcolsep}{3.0pt}
\renewcommand{\arraystretch}{1.08}
\begin{tabular*}{\columnwidth}{@{\extracolsep{\fill}}lcccccc@{}}
\toprule
\textbf{Model} & \textbf{Tel.} & \textbf{Proc.} & \textbf{Comp.}
& \textbf{Rel.} & \textbf{Final} & \textbf{Code} \\
\midrule
Llama    & 64.50 & 99.25 & 96.75 & 50.75 & 51.50 & 61.75 \\
Qwen     & 78.36 & 98.68 & 97.36 & 44.59 & 44.06 & 76.25 \\
GLM      & 65.41 & 99.71 & 99.42 & 65.70 & 65.41 & 62.50 \\
DeepSeek & 68.82 & 85.39 & 98.03 & 45.51 & 42.42 & 58.43 \\
\bottomrule
\end{tabular*}
\caption{
Sensitivity of the structured intermediate-output evaluation under prompt
variant~1. Abbreviations follow
Table~\ref{tab:intermediate_sensitivity_original}. All values are percentages.
}
\label{tab:intermediate_sensitivity_v1}
\end{table}

\begin{table}[t]
\centering

\scriptsize
\setlength{\tabcolsep}{3.0pt}
\renewcommand{\arraystretch}{1.08}
\begin{tabular*}{\columnwidth}{@{\extracolsep{\fill}}lcccccc@{}}
\toprule
\textbf{Model} & \textbf{Tel.} & \textbf{Proc.} & \textbf{Comp.}
& \textbf{Rel.} & \textbf{Final} & \textbf{Code} \\
\midrule
Llama    & 69.17 & 99.50 & 78.70 & 54.64 & 55.14 & 58.40 \\
Qwen     & 76.17 & 89.38 & 100.00 & 38.60 & 38.60 & 65.54 \\
GLM      & 73.16 & 100.00 & 98.23 & 67.51 & 67.51 & 72.41 \\
DeepSeek & 64.33 & 91.52 & 92.40 & 51.46 & 51.46 & 60.82 \\
\bottomrule
\end{tabular*}
\caption{
Sensitivity of the structured intermediate-output evaluation under prompt
variant~2. Abbreviations follow
Table~\ref{tab:intermediate_sensitivity_original}. All values are percentages.
}
\label{tab:intermediate_sensitivity_v2}
\end{table}
\section{Prompts}

\newtcolorbox{promptbox}{
    enhanced,
    breakable,
    colback=gray!5,
    colframe=gray!70,
    boxrule=0.5pt,
    arc=2mm,
    left=2mm,
    right=2mm,
    top=1mm,
    bottom=1mm,
    before skip=6pt,
    after skip=8pt,
    fontupper=\small
}

\label{sec:prompt_design}

We design the prompts to distinguish changes in the model's final label
preference from improvements in its event-semantic reasoning. All prompts
instruct the model to rely only on the premise and to avoid adding
commonsense assumptions. Unless otherwise specified, the model predicts one
of three NLI labels: \textsc{True}, \textsc{False}, or \textsc{Unknown}.
\textsc{True} indicates that the hypothesis is necessarily true given the
premise, \textsc{False} indicates that it is contradicted by the premise,
and \textsc{Unknown} indicates that it is neither entailed nor contradicted.

\subsection{Baseline Prompts}
\label{sec:baseline_prompts}
\paragraph{Zero-Shot Prompt}
The zero-shot prompt provides only the standard definition of three-way NLI: the hypothesis should be labeled as entailed, contradicted, or unknown depending on whether it is necessarily true, necessarily false, or unresolved given the premise.
This information defines the final classification task, but it does not provide any guidance about the event-semantic distinctions required to solve the benchmark. 
The zero-shot prompt therefore requires the model to recover the entire semantic analysis implicitly before producing a single label.

\paragraph{DAP Prompt}







The original study presents the DAP prompt as providing a definition of the imperfective paradox. In practice, however, the prompt states only two intended inference patterns:

\begin{quote}
For accomplishments or goal-oriented actions, the progressive form does not imply completion.

For activities, the process implies that the action occurred.
\end{quote}

These statements describe how the two aspectual classes are expected to behave, but they do not explain how the model should determine whether the relevant predicate is an accomplishment or an activity. The prompt directly invokes the terms \textit{accomplishment}, \textit{goal-oriented action}, and \textit{activity} without providing operational criteria for distinguishing among them.
This omission is consequential because lexical aspect is determined compositionally by the complete predicate rather than by the isolated verb. 
The expression \textit{goal-oriented action} is also potentially misleading. Many activities are performed in pursuit of an external goal, even though the linguistic predicate does not itself encode an inherent endpoint.
Most importantly, the prompt omits the central truth-conditional distinction underlying the imperfective paradox. For an atelic simple-past predicate, evidence that the relevant process occurred is normally sufficient. For a telic simple-past predicate, by contrast, occurrence alone is insufficient; the event must reach the culmination encoded by the predicate. Thus, the critical distinction is not merely that the progressive fails to entail completion for accomplishments, but that the two classes impose different standards for validating their corresponding simple-past forms.

\paragraph{Chain-of-Thought Prompt}

The original chain-of-thought prompt also does not fully instantiate the reasoning process required by the task.
\begin{quote}
First, analyze the temporal status of the event in the premise. Does the action have a defined endpoint? Was it completed?
\end{quote}

In particular, it does not explicitly require the model to classify the complete predicate as telic or atelic before reasoning about completion.
Instead, it asks the model to reason using terms such as \textit{temporal status} and \textit{defined endpoint} without providing precise definitions. The phrase \textit{temporal status} is underspecified and may be interpreted as whether the event occurs in the past, present, or future, whether it is ongoing or terminated, or whether it has culminated. These are related but distinct properties.
Likewise, the phrase \textit{defined endpoint} may be misleading. A model may interpret it as the time at which an event happens to stop rather than as an inherent culmination encoded by the predicate.
The prompt therefore asks the model to reason through several semantic notions without defining them or explicitly enforcing the order in which they should be analyzed. Merely requesting chain-of-thought reasoning does not guarantee that the model will perform the theoretically relevant intermediate steps.
\paragraph{Zero-shot.}

\begin{promptbox}
\textbf{System:}\\
You are a strict logician. Your task is to determine if a Hypothesis is
necessarily true given a Premise.

If the Hypothesis MUST be true based only on the Premise, output
``True''. If the Hypothesis is contradicted by the Premise, output
``False''. If the Hypothesis might be true but is not explicitly
guaranteed by the Premise, output ``Unknown''.

Do not use common-sense assumptions. Only use the text provided.

\textbf{User:}\\
Premise: \{premise\}\\
Hypothesis: \{hypothesis\}\\
Respond with only one option: True, False, or Unknown.
\end{promptbox}

\paragraph{Definition-aware prompting (DAP).}

\begin{promptbox}
\textbf{System:}\\
You are a strict logician. Determine whether the Hypothesis is necessarily
true given the Premise.

Use the following linguistic rules:

\begin{itemize}
    \item For accomplishments or goal-oriented actions, the progressive form
    does not imply completion.
    \item For activities, evidence that the process occurred implies that the
    action occurred.
\end{itemize}

If the Hypothesis must be true, output ``True''. If it is contradicted,
output ``False''. If it may be true but is not guaranteed, output
``Unknown''.

Do not use common-sense assumptions. Use only the Premise.

\textbf{User:}\\
Premise: \{premise\}\\
Hypothesis: \{hypothesis\}\\
Respond with only one option: True, False, or Unknown.
\end{promptbox}

\paragraph{Chain-of-thought prompting (CoT).}

\begin{promptbox}
\textbf{System:}\\
You are a strict logician. Determine whether the Hypothesis is necessarily
true given the Premise.

First, analyze the temporal status of the event in the Premise. Does the
action have a defined endpoint? Was it completed? Then provide the final
label.

Do not add information not stated in the Premise.

\textbf{User:}\\
Premise: \{premise\}\\
Hypothesis: \{hypothesis\}\\
Return only the following JSON object:
\begin{flushleft}
\ttfamily
\{\\
\hspace*{1.0em}"reasoning": "...",\\
\hspace*{1.0em}"label": "True/False/Unknown"\\
\}
\end{flushleft}
\end{promptbox}

\paragraph{Counterfactual prompting.}

\begin{promptbox}
\textbf{System:}\\
You are a strict logician. Determine whether the Hypothesis follows from the
Premise.

Before predicting the label, consider whether the event described in the
Premise could have been interrupted or could have stopped before reaching
the endpoint required by the Hypothesis. If such a non-endpoint continuation
is compatible with the Premise, do not assume completion.

Use only information supplied by the Premise.

\textbf{User:}\\
Premise: \{premise\}\\
Hypothesis: \{hypothesis\}\\
Return only the following JSON object:
\begin{flushleft}
\ttfamily
\{\\
\hspace*{1.0em}"counterfactual\_analysis": "...",\\
\hspace*{1.0em}"label": "True/False/Unknown"\\
\}
\end{flushleft}
\end{promptbox}

\subsection{Revised Event-Semantic Prompt}
\label{sec:revised_prompt}

The baseline prompts do not fully specify how predicate type, process
occurrence, and culmination jointly determine the truth of a simple-past
hypothesis. We therefore construct DAPCoT, which explicitly decomposes the
decision into the following steps:

\begin{enumerate}
    \item extract and classify the complete predicate;
    \item determine whether its event process occurred;
    \item identify the endpoint encoded by the predicate, if any;
    \item determine whether culmination or non-culmination is established;
    \item apply the truth conditions appropriate to the predicate class; and
    \item map the resulting semantic judgment to an NLI label.
\end{enumerate}

The complete DAPCoT prompt is given below.
\begin{promptbox}

\textbf{System:}\\

You are a strict logician. Determine whether a Hypothesis is ``True'',
``False'', or ``Unknown'' given a Premise.

If the Hypothesis MUST be true based only on the Premise, output ``True''.

If the Hypothesis is contradicted by the Premise, output ``False''.

If the Hypothesis might be true but is not explicitly guaranteed, output
``Unknown''.

\textbf{User:}\\

Premise: \{premise\}\\

Hypothesis: \{hypothesis\}

\textbf{Important linguistic rule:}

\begin{itemize}

    \item \textbf{Accomplishments/goal-oriented actions:} The predicate
    includes an inherent endpoint, such as a bounded path, fixed quantity,
    created object, required state, or required result, as in
    ``build a house.''

    \item \textbf{Activity actions:} The predicate describes an ongoing
    process with no inherent required endpoint or result, as in ``run.''

    \item Judge the complete predicate, including its object and complements.
    Use only endpoints explicitly encoded by the predicate. Do not infer an
    endpoint from common sense, typical purposes or outcomes, the event merely
    stopping, participants, objects, locations, stimuli, audiences, or
    reactions.

    \item For accomplishments or goal-oriented actions, the progressive form
    does not imply completion. A simple-past hypothesis is true only if the
    action was completed.

    \item For activity actions, the process implies that the action occurred.
    A simple-past hypothesis is true as long as the action occurred, and
    completion is not required.

\end{itemize}

Let's think step by step.

\begin{enumerate}

    \item Extract the action in the Hypothesis.

    \item Find its inherent endpoint. If no inherent endpoint is encoded,
    state that explicitly; do not treat the situation itself as an inherent
    endpoint.

    \item Classify the action.

    \item Apply the appropriate rule.

\end{enumerate}

Use only the facts stated in the Premise. Do not use common-sense assumptions.

Do not output code, pseudocode, Markdown, or code fences.

Return exactly one valid JSON object and nothing else:

\begin{flushleft}

\ttfamily

\{\{\\

\hspace*{1.0em}"reasoning": "<brief reasoning>",\\

\hspace*{1.0em}"label": "True/False/Unknown"\\

\}\}

\end{flushleft}

\end{promptbox}

\subsection{Prompt Ablations}
\label{sec:prompt_ablations}

We construct three ablations to provide more information on which parts of DAPCoT account for
its performance.

\paragraph{DAPCoT-n.}
This variant removes the following instruction:

\begin{promptbox}
the progressive form does not imply completion.

the process implies that the action occurred.
\end{promptbox}

The ablation tests whether models can independently distinguish process
occurrence from culmination.

\paragraph{DAPCoT-d.}
This variant removes the
operational definitions of activities and accomplishments:

\begin{promptbox}
Accomplishments/goal-oriented actions: The predicate includes an inherent endpoint, such as
a bounded path, fixed quantity, created object, required state, or required result, as in “build a
house.”
Activity actions: The predicate describes an ongoing process with no inherent required endpoint
or result, as in “run.”
\end{promptbox}

This ablation tests the contribution of explicit predicate-class
definitions.

\paragraph{DAPCoT-p.}
This variant removes the truth-conditional distinction between the
two predicate classes:

\begin{promptbox}
For an activity predicate, occurrence of the relevant process is sufficient
for the corresponding simple-past statement. For an accomplishment
predicate, occurrence alone is insufficient; the endpoint encoded by the
predicate must be reached.
\end{promptbox}

DAPCoT-p therefore retains a general request for reasoning but does not tell
the model how its intermediate semantic judgments determine the truth of the
simple-past hypothesis.

\subsection{Diagnostic Prompts}
\label{sec:diagnostic_prompts}

\paragraph{Endpoint probe.}
The endpoint probe directly tests whether the premise guarantees culmination.

\begin{promptbox}
\textbf{System:}\\
You are a strict logician. Based only on the Premise, determine whether the
endpoint required by the event in the Hypothesis is reached.

Answer ``True'' if the Premise guarantees that the endpoint was reached.
Answer ``False'' if the Premise guarantees that the endpoint was not
reached. Answer ``Unknown'' if the Premise does not determine whether the
endpoint was reached.

Absence of evidence for completion is not evidence of non-completion. Do not
use common-sense assumptions.

\textbf{User:}\\
Premise: \{premise\}\\
Hypothesis: \{hypothesis\}\\
Respond with only one option: True, False, or Unknown.
\end{promptbox}

\paragraph{Event-status probe.}
To distinguish occurrence from culmination, we ask models to classify the
event into one of three states.

\begin{promptbox}
\textbf{System:}\\
You are a strict logician. Based solely on the Premise, determine the status
of the event described in the Hypothesis.

Choose exactly one option:

A. The event occurred, but the Premise does not establish whether it was
completed or interrupted.

B. The event occurred, but the Premise explicitly indicates that the event
process was interrupted or prematurely terminated.

C. The Premise establishes that the event was completed.

Do not make common-sense assumptions. Base your answer only on information
explicitly provided by the Premise.

\textbf{User:}\\
Premise: \{premise\}\\
Hypothesis: \{hypothesis\}\\
Respond with exactly one letter: A, B, or C.
\end{promptbox}

For the reasoning condition, the model receives the same options together
with the following instruction:

\begin{promptbox}
Reason step by step. First determine whether the event occurred. Then
determine whether the Premise establishes completion, explicitly indicates
interruption or premature termination, or leaves the status unresolved.
Return:
\begin{flushleft}
\ttfamily
\{\\
\hspace*{1.0em}"reasoning": "...",\\
\hspace*{1.0em}"final\_answer": "A/B/C"\\
\}
\end{flushleft}
\end{promptbox}

\paragraph{Intermediate-output probe.}
Finally, we elicit the intermediate decisions and final prediction jointly.
\begin{promptbox}

\textbf{System:}\\

You are a strict logician. Your task is to determine if a Hypothesis is
necessarily true given a Premise.

If the Hypothesis MUST be true based only on the Premise, output ``True''.

If the Hypothesis is contradicted by the Premise, output ``False''.

If the Hypothesis might be true but is not explicitly guaranteed by the
Premise, output ``Unknown''.

Do not use common-sense assumptions. Only use the text provided.

\textbf{User:}\\

Premise: \{premise\}\\

Hypothesis: \{hypothesis\}

Before giving the final answer, reason and provide the following intermediate
labels based only on the Premise:

\begin{enumerate}

    \item Based on the complete predicate in the Hypothesis, classify it as
    an accomplishment/goal-oriented action or an activity.

    \item Based only on the Premise, does the Premise establish that the
    event process occurred?

    \item Based only on the Premise, does the Premise guarantee that the
    event was completed?

    \item Based only on the Premise, does an interruption described in the
    Premise rule out completion?

    \item Based only on the Premise, what is the logical relation of the
    Hypothesis to the Premise?

\end{enumerate}

Do not add common-sense assumptions or information not stated in the Premise.
Base the reasoning, every intermediate label, and the final answer only on the
Premise.

Please respond with ONLY the following JSON object, replacing each value with
one of its listed options:

\begin{flushleft}

\ttfamily

\{\\

\hspace*{1.0em}"reasoning": "...",\\

\hspace*{1.0em}"predicate\_telicity":
"accomplishments/goal-oriented actions"/"activities",\\

\hspace*{1.0em}"process\_occurred": "Yes"/"No",\\

\hspace*{1.0em}"completion\_is\_guaranteed": "Yes"/"No",\\

\hspace*{1.0em}"interruption\_rules\_out\_completion": "Yes"/"No",\\

\hspace*{1.0em}"nli\_relation":
"Entailment"/"Contradiction"/"Neutral",\\

\hspace*{1.0em}"final\_answer": "True"/"False"/"Unknown"\\

\}

\end{flushleft}

\end{promptbox}

\section{Implementation Details}

\label{sec:implementation_details}

\paragraph{Models and inference.}
We evaluate four instruction-tuned language models at the 7B--9B scale:
Llama-3.1-8B-Instruct, Qwen2.5-7B-Instruct, GLM-4-9B-0414, and
DeepSeek-R1-Distill-Qwen-7B, with temperature set to
zero and a maximum output length of 2,048 tokens.  DeepSeek-R1-7B is reasoning-oriented and typically generates an explicit reasoning trace before producing its final answer. For DeepSeek-R1-7B, both settings include the model's native reasoning process. The distinction lies in the final answer format. In the \textit{Label only} setting, the final answer contains only the predicted label. In the \textit{Reasoning + label} setting, the model is additionally instructed to provide a task-specific justification in the final answer before the label. This justification is separate from the model's native reasoning trace.

\paragraph{Prompting baselines.}
We evaluate zero-shot, definition-aware prompting (DAP), chain-of-thought
(CoT), and counterfactual prompting. All prompts use the same three-way NLI
label space: \textsc{True} if the hypothesis is entailed by the premise,
\textsc{False} if it is contradicted, and \textsc{Unknown} if its truth is
not determined. The zero-shot condition provides only these label
definitions. DAP additionally states the intended inference patterns for
activities and accomplishments. CoT asks the model to analyze the temporal
status and completion of the event before predicting a label. The
counterfactual prompt asks the model to consider ways in which the event
could fail to reach its endpoint. For Llama, Qwen, and GLM, the
no-reasoning setting requests only the final label, whereas the reasoning
setting requests an explicit analysis followed by a structured final
answer. DeepSeek-R1 is evaluated using its native reasoning behavior, with
either a label-only or a reasoning-plus-label output instruction as
appropriate.

\paragraph{Completion and event-status diagnostics.}
For the direct completion probe, given the premise and hypothesis, models judge
whether the endpoint required by the hypothesis is established by the
premise, using the labels \textsc{True}, \textsc{False}, and
\textsc{Unknown}. We compare a direct-answer prompt with a reasoning prompt
that first requires the model to assess the endpoint evidence. Accuracy is
computed against \textsc{Unknown}, while NT is the proportion of predictions
that are not \textsc{True}.

For the group-wise event-status diagnostic, we sample twenty unambiguous, manually verified examples from each
of Groups A--D. Models choose among three states: (i) the event occurred, but
its completion status is unresolved, (ii) the event occurred and was
explicitly interrupted or prematurely terminated, and (iii) the event was
completed. The gold state is interruption for Groups A and B and unresolved
status for Groups C and D. We report classification accuracy and the
non-completion-class rate (NC Rate), calculated as the proportion of
predictions not assigned to the completed-event class. The same examples
are evaluated under direct-answer and step-by-step reasoning prompts.

\paragraph{Occurrence-to-completion intervention.}
To determine when a model begins to accept a telic simple-past hypothesis,
we construct five premise variants that provide progressively different
information about the same event: intention, occurrence, interruption,
explicit non-completion, and completion. Each condition contains twenty
examples. The hypothesis remains fixed across the variants, allowing the
decision change to be attributed to the added event information. We report
the percentage of examples for which the model makes the correct prediction at each stage. We additionally use a
three-way contrast among an occurred-but-unresolved event, an interrupted
event, and a completed event to separate representations of event
occurrence from representations of culmination.

\paragraph{Revised prompts and ablations.}
Our revised DAPCoT prompt explicitly decomposes the task into predicate
classification, process verification, endpoint identification, culmination
and non-culmination verification, semantic-rule application, and final NLI
label alignment. It defines activities as predicates without an inherent
endpoint and accomplishments as predicates whose complete verb phrase
encodes a required endpoint or result. It further states that occurrence is
normally sufficient for an atelic simple-past predicate, whereas culmination
is required for a telic simple-past predicate. The prompt instructs models
to classify the complete predicate, including its object and complements,
and not to infer endpoints from typical goals, world knowledge, event
termination, or the mere presence of an object or location.

We evaluate three ablations. DAPCoT-n removes the statement that the
progressive establishes occurrence without guaranteeing completion.
DAPCoT-d removes the operational definitions of activities and
accomplishments. DAPCoT-p additionally removes the distinction between the
truth conditions of their simple-past forms. These configurations use the
same examples, decoding parameters, and output-label normalization as the
main prompting experiments.

\paragraph{Intermediate-output evaluation.}
The model produces its intermediate semantic
decisions and final NLI prediction. The evaluated
fields include predicate telicity, process occurrence, completion,
interruption, the predicted NLI relation, and the final
\textsc{True}/\textsc{False}/\textsc{Unknown} answer. We report field-level
accuracy, final-answer accuracy, and the percentage of responses with
all correct fields. In addition to the model's own final prediction, we
derive a code-mapped prediction by deterministically applying the intended
semantic rules to the extracted intermediate variables. This comparison
separates errors in intermediate semantic discrimination from errors in
mapping otherwise appropriate semantic judgments to an NLI label.

\label{de}

\end{document}